\pdfoutput=1

\documentclass[11pt]{article}

\usepackage{acl}

\usepackage{times}
\usepackage{latexsym}
\usepackage{fdsymbol}
\usepackage[T1]{fontenc}

\usepackage[utf8]{inputenc}

\usepackage{microtype}

\usepackage{hyperref}
\usepackage{booktabs}
\usepackage{graphicx}
\graphicspath{{./imgs/}}
\usepackage{footmisc}
\usepackage{microtype}

\usepackage{caption}
\usepackage{subcaption}

\usepackage{amsmath}
\usepackage{amsfonts,bm}
\usepackage{xspace}

\newcommand{\red}[1]{\textcolor{red}{#1}}
\newcommand{\blue}[1]{\textcolor{blue}{#1}}

\newcommand{\ie}{{\em i.e.,}\xspace}
\newcommand{\eg}{{\em e.g.,}\xspace}
\newcommand{\wrt}{\emph{w.r.t.}\xspace}

\newcommand{\Ni}{({\em i})~}
\newcommand{\Nii}{({\em ii})~}
\newcommand{\Niii}{({\em iii})~}

\definecolor{mypink3}{cmyk}{0, 0.7808, 0.4429, 0.1412}

\makeatletter   
\newcommand{\sveryshortarrow}[1][3pt]{\mathrel{%
    \vcenter{\hbox{\rule[-.5\fontdimen8\scriptfont3]
               {\scriptratio\dimexpr#1\relax}{\fontdimen8\scriptfont3}}}%
   \mkern-4mu\hbox{\let\f@size\sf@size\usefont{U}{lasy}{m}{n}\symbol{41}}}}
\makeatother

\def\eqref#1{equation~\ref{#1}}

\def\1{\bm{1}}

\def\m1{{\bm{1}}}

\DeclareMathAlphabet{\mathsfit}{\encodingdefault}{\sfdefault}{m}{sl}
\SetMathAlphabet{\mathsfit}{bold}{\encodingdefault}{\sfdefault}{bx}{n}

\def\gD{{\mathcal{D}}}

\def\gL{{\mathcal{L}}}

\usepackage[nameinlink]{cleveref}
\crefformat{section}{\S#2#1#3} 
\crefname{algorithm}{Alg.}{Algs.}
\crefformat{subsection}{\S#2#1#3}
\Crefname{equation}{Eq.}{Eqs.}
\Crefname{figure}{Fig.}{Figs.}

\usepackage[colorinlistoftodos,prependcaption,textsize=tiny]{todonotes}

\usepackage{soul}

\newcommand{\change}[1]{{\leavevmode\color{black}#1}}
\title{Chart-to-Text: A Large-Scale Benchmark for Chart Summarization}

\author{
Shankar Kantharaj$^{\clubsuit}$\thanks{\ \ Equal contribution. Listing order is based on the alphabetical ordering of author surnames.}, \  Rixie Tiffany Ko Leong$^{\vardiamondsuit}$\footnotemark[1], \ Xiang Lin$^\vardiamondsuit$\footnotemark[1], \ Ahmed Masry$^{\clubsuit}$\footnotemark[1]\\ \textbf{Megh Thakkar$^{\vardiamondsuit}$\footnotemark[1], \ Enamul Hoque$^{\clubsuit}$,  \ Shafiq Joty$^{\vardiamondsuit\spadesuit}$}\\
$^\clubsuit$York University, Canada, 
$^\vardiamondsuit$Nanyang Technological University, Singapore\\
$^\spadesuit$Salesforce Research Asia, Singapore\\
\{shankark, masry20, enamulh\}@yorku.ca\\
\{rleong007, linx0057, srjoty\}@ntu.edu.sg \\
megh.1211@gmail.com \\
}

\begin{document}
\maketitle

\begin{abstract}
Charts are commonly used for exploring data and communicating insights. Generating natural language summaries from charts can be very helpful for people in inferring key insights that would otherwise require a lot of cognitive and perceptual efforts. 
We present Chart-to-text, a large-scale benchmark with two datasets and a total of 44,096 charts covering a wide range of topics and chart types. We explain the dataset construction process and analyze the datasets. We also introduce a number of state-of-the-art neural models as baselines that utilize image captioning and data-to-text generation techniques to tackle two problem variations: one assumes the underlying data table of the chart is available while the other needs to extract data from chart images. Our analysis with automatic and human evaluation shows that while our best models usually generate fluent summaries and yield reasonable BLEU scores, they also suffer from  hallucinations and factual errors as well as difficulties in correctly explaining complex patterns and trends in charts.

\end{abstract}

\section{Introduction} \label{sec:intro}

Data visualizations such as bar charts, line charts, and pie charts are very popular for presenting quantitative data. Often people use such charts to get important insights from data and make informed decisions. However, it is well-known that inferring key insights from the charts can be quite challenging and time-consuming, as it may require a lot of cognitive and perceptual efforts~\cite{perez, whitaker}. 

Automatic chart summarization is a task where the goal is to explain a chart and summarize key takeaways from it in natural language. Chart summarization has several key benefits and potential applications. First, chart summaries can help people  identify key insights from charts that they might have missed otherwise. In a study on a chart corpus, \citet{carberry2006information} found that chart authors often failed to convey key insights from charts in their corresponding textual captions. Thus, automatic summarization could help authors write effective reports and articles on data facts by suggesting explanatory texts. Similarly, readers could benefit from such summaries, as studies have found that captions help readers find important points by explaining visually prominent features in charts~\citep{kim2021towards}.
Chart summarization offers another important benefit of making charts more accessible to people who are visually impaired since they can use screen readers to understand what is being presented in the chart~\cite{Ferres-accessibility-2013}. Finally, the generated summaries can be leveraged for indexing documents containing charts to improve information retrieval algorithms~\cite{li2013towards}.

\begin{figure}[t] 
\linespread{0.5}\selectfont\centering 
\scriptsize
\scalebox{0.99}{\begin{tabular}{@{}p{1\linewidth}} 
        \includegraphics[width=7.8cm]{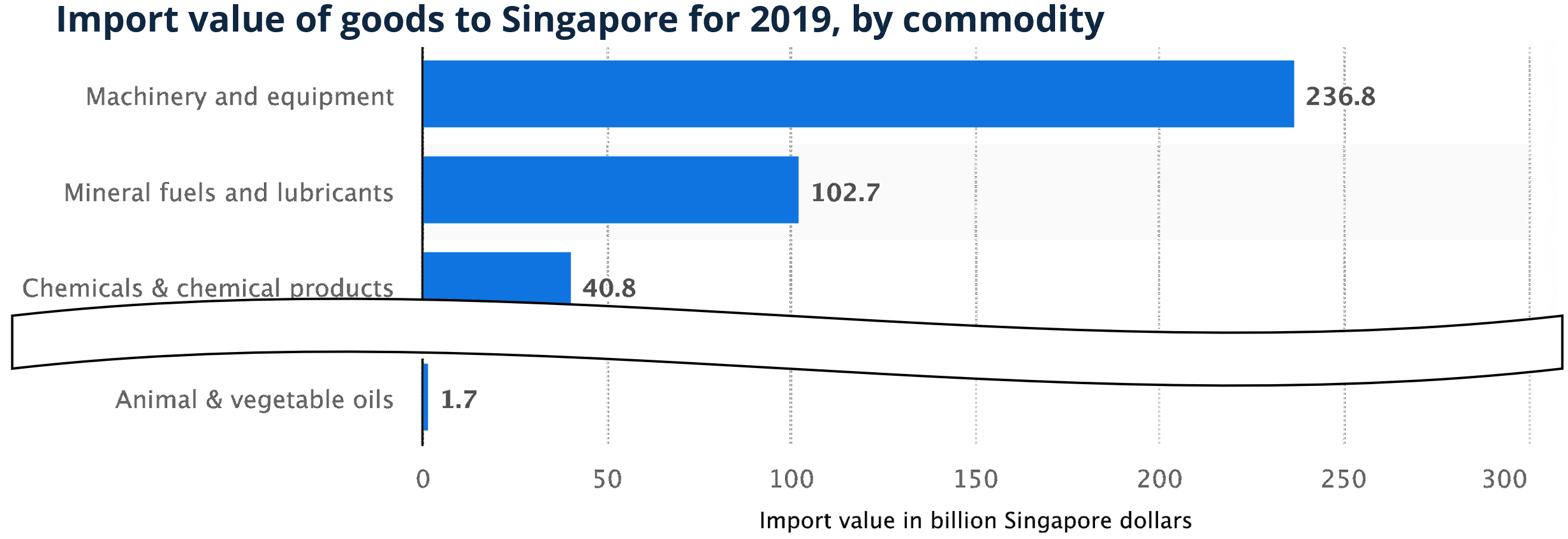} \\ [+0.5em]
        { 
        \textbf{Gold:} In 2019, Singapore imported approximately 236.8 billion Singapore dollars worth of machinery and equipment, making it the country's largest import commodity by value. This was followed by the import of mineral fuels and lubricants, valued at 102.7 billion Singapore dollars.}
      \\ [+0.3em]
        {\textbf{TAB-T5:} Machinery and equipment was the most valuable commodity for Singapore in 2019, with an import value of 236.8 billion Singapore dollars. Mineral fuels and lubricants were the second most valuable commodity for Singapore, with an import value of 102.7 billion Singapore dollars.}
    \end{tabular}
    }
    \caption{An example chart-summary pair from our Benchmark and the output from one of the best models (TAB-T5).}
    \label{tab:example}
\end{figure}

Despite its numerous benefits and applications, the chart summarization problem  has not received much attention in the NLP community.  Early approaches relied on template-based text generation methods that combine statistical techniques and planning-based architecture~\cite{reiter2007architecture} to generate captions from bar and line charts~\cite{fasciano1996postgraphe, mittal-etal-1998-describing, green2004autobrief,  demir-etal-2012-summarizing}. Recently, researchers considered data-driven neural models for describing tabular data~\cite{mei2016talk, gong-etal-2019-enhanced}. 
\change{However, compared to tables, charts serve a different communication goal, and so is the chart-to-text problem. Unlike tables which simply list raw data, 
charts create visual representation of data that can draw a reader’s attention to various prominent features such as trends and outliers~\cite{kim2021towards}.  For example, a line chart may depict an important trend whereas a scatterplot may visually communicate correlations and outliers. 
Existing table-to-text approaches are not designed to explain such visually salient chart features in summaries.}

There are two main impediments to addressing the chart summarization task. First, the lack of large-scale datasets makes it difficult to solve the  task using data-driven neural models. Second, there are no strong baselines that  utilize the latest advances in neural text generation tasks. \citet{obeid} made an initial attempt to address this problem with a dataset and a model that utilizes a Transformer \cite{Vaswani2017attn}  architecture. 
However, their dataset was built by collecting a small set of charts (8,305) from a single source covering only two types of charts (bar and line). Also, their approach does not exploit the recent advances in large-scale language model pretraining, which has been shown to be very beneficial for many vision and language tasks \cite{devlin-etal-2019-bert,pmlr-v139-touvron21a}. To our knowledge, there is no large-scale benchmark with a wider range of topics from multiple sources, covering many different chart types, and with models that employ large-scale pretraining.

In this work, we present a large-scale benchmark for chart-to-text with two datasets consisting of 44,096 charts  covering a broad range of topics and a variety of chart types. We introduce two variations of the problem. The first variation assumes that the underlying data table of a chart is available, while the other introduces a more challenging and realistic scenario by assuming that the chart is in image format and the underlying table is not available. These two problem scenarios motivated us to adapt a variety of state-of-the-art models that combine computer vision and natural language generation techniques as strong baselines; see \Cref{tab:example} for a sample model output.

Our primary contributions are: \Ni a new large-scale benchmark 
covering a wide range of topics and chart types; \Nii a set of state-of-the-art neural models 
which can act as a starting point for other researchers to expand and improve upon; and \Niii a series of automatic and human evaluations as well as in-depth qualitative analysis to identify further challenges.
Our code and benchmark datasets are publicly available at
\href{https://github.com/vis-nlp/Chart-to-text}{https://github.com/vis-nlp/Chart-to-text}.

\section{Related Work}

\paragraph{Chart Summarization}

Early work 
\cite{mittal-etal-1998-describing,Ferres-accessibility-2013} followed a planning-based architecture ~\cite{reiter2007architecture} and used templates to generate texts. 
These systems only describe how to read the chart rather than explain key insights conveyed by the chart. 
Recently, commercial systems such as  Quill and Wordsmith\footnote{Narrative Science \href{https://narrativescience.com/quill/}{Quill}; Automated Insights \href{https://automatedinsights.com/wordsmith}{Wordsmith}}
as well as research prototypes, \eg\ \cite{cui2019datasite} and  \cite{srinivasan2018augmenting}  
computed statistics  (\eg\ extrema, outliers) to present facts from a dataset. \citet{demir-etal-2012-summarizing} also compute statistics to generates bar chart summaries in a bottom–up manner to simultaneously construct the discourse and sentence structures. Recently, \citet{DBLP:journals/corr/abs-1906-02850} used the ResNet \cite{he2016deep} to encode the chart image and an LSTM decoder to create the caption. 

A key limitation of the above bodies of work is that sentences are generated using predefined templates, which may lack generality and offer little variation in terms of reported insights, grammatical styles and lexical choices compared to data-driven models. Moving beyond template-based summaries, \citet{obeid} adapted a transformer-based  model on a dataset of 8,305 
charts,  while \citet{Andrea-carenini} applied an LSTM based encoder-decoder model on a dataset of 306 
chart summaries. Both studies used much smaller datasets and did not consider the computer vision aspects of the problem. \change{~\citet{hsu2021scicap} recently use a CNN+LSTM based image captioning model for scientific figure captioning.} In contrast, \change{we focus on the generic chart-to-text problem and} train several 
 neural models that combine computer vision and data2text generation.

\paragraph{Data2text Generation}
Data2text models generate a descriptive summary for a table of records. They have been used for various domain-specific tasks such as summarizing sports data~\cite{barzilay-lapata-2005-collective,  
wiseman2017challenges}, weather-forecast data
~\cite{reiter2005choosing}, recipe generation~\cite{yang2017reference} and biography generation~\cite{lebret-etal-2016-neural} as well as open-domain tasks~\cite{parikh2020totto, chen2020logical}. Recent methods have primarily used an LSTM-based encoder-decoder architecture~\cite{mei2016talk, lebret-etal-2016-neural, wiseman2017challenges}.  \citet{gong-etal-2019-enhanced} found that transformers \cite{Vaswani2017attn} yielded more fluent and coherent outputs compared to their LSTM counterparts. Others focused on \change{controlling the structure of the summary using a planning approach~\cite{su2021plangen} as well as generating facts by preforming logical inference over the given table~\cite{chen2020logical,chen-etal-2020-logic2text}.} 

\paragraph{Image Captioning}
There has been swift progress 
in image captioning largely due to the availability of large-scale datasets \cite{agrawal2019nocaps, chen2015microsoft}.  ~\citet{zhang2021vinvl} developed an 
object detection model to summarize objects in images while 
\citet{sidorov2020textcaps} utilized texts extracted from images using OCR  to generate captions. 
Unlike images with real-world objects and scenes, charts have marks (\eg\ bars,  lines) that map quantitative data. This makes the chart-to-text problem different from image captioning.

\section{Chart-to-text Datasets}

\change{After searching through various sources including news sites, textbooks, and websites containing data facts}, we found two suitable sources with sufficiently large numbers \change{and varieties} of charts with textual descriptions as we describe below.

\subsection{Data Collection}

\paragraph{$\bullet$ Statista} Statista ({\href{https://www.statista.com}{statista.com}}) is an online platform that regularly publishes charts on a wide range of topics including economics, market and opinion research. 
We crawled 34,810 publicly accessible webpages in December 2020, yielding a total of 34,811 charts. 
For each chart, we took a screenshot of the chart image, downloaded the data table, the title, axis labels and the human-written 
descriptions about the chart. We classified the charts into two groups based on the number of columns in their underlying data tables: Data tables of \emph{simple} charts  have only two columns, whereas \emph{complex} charts involve at least three columns (\eg\ stacked or group bar charts, line charts with multiple lines).

\paragraph{$\bullet$ Pew} The Pew Research ({\href{https://www.pewresearch.org}{pewresearch.org}}) publishes data-driven articles about social issues, public opinion and demographic trends. The articles are often accompanied by multiple charts along with high-quality descriptions written by professional editors. We scraped 3,999 publicly accessible pages in January 2021, which gave a total of 9,285 charts. Unlike Statista, the Pew reports 
do not provide the underlying data tables for most of the charts. Among 9,285 charts, only 143 have 
underlying data tables. For each chart, we downloaded the chart image, the surrounding paragraphs and the alternative text associated with the image (using the \texttt{alt} attribute), if it was available. Like a title, the \texttt{alt} text often gives a very short chart description. Finally, we classified the charts into \emph{simple} and \emph{complex} manually since 
underlying data tables were unavailable. 

\subsection{Data Annotation} \label{subsec:data-annot}

\change{Below we describe two main steps of the data annotation process for each chart:  \Ni identify the relevant summary, and \Nii extract data. Additional details of these steps are provided in \Cref{app:annotation-details}}.
\paragraph{$\bullet$ Statista} We chose  the first part of the text 
(from the chart icon to the next heading) as the chart summary. 
This is based on the observation that the first part provides a succinct summary of the chart while the remaining parts often contain background information (\eg\ the history of a company).

Extracting data from the Statista charts was relatively straightforward as the underlying data tables were available. However,
most charts (32,660 out of 34,811) did not provide x-axis labels. To assign representative labels for them, we first used regular expressions on the cell values of such a column to see if it represents common entities (\eg \emph{year}, \emph{location}). Still, there were 7,170 missing labels remaining. We then applied the Wikidata knowledge base~\cite{Wikidata} to automatically derive an entity type label based on the data values plotted on x-axis. However, sometimes the resulting labels were too generic (\eg \emph{human}, \emph{business}). Hence, we manually annotated each label by either accepting the entity type label, if it
represents the x-axis accurately, or entering a more specific name.

\begin{figure}[t]
    \centering
    \includegraphics[width=0.49\textwidth]{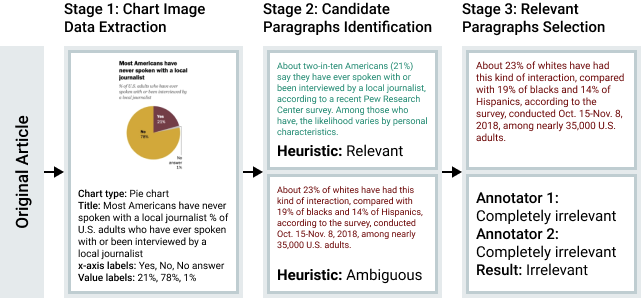}
    \caption{Stages of the Pew dataset construction process.}
    \label{pew-annot-process}
\end{figure}

\paragraph{$\bullet$ Pew} The annotation 
for Pew was more challenging as often a webpage  contains many charts and
paragraphs do not explicitly refer to their relevant chart.
\change{Also, most charts did not have underlying data tables.}
To address these challenges, we construct the dataset in three stages
(\Cref{pew-annot-process}).

\vspace{0.5em}
\noindent \Ni \textit{Data extraction from chart images: } We first extracted the text from the charts using CRAFT \cite{craftrecognition,craftdetection},  a state-of-the-art OCR model. We then  extracted the bounding boxes of the detected texts to extract geometric features (\eg normalized width and height of the text) and used them to train a gradient boosting classifier that categorizes the recognized text into one of the  following categories: title, axis labels, legends, and data labels. Since the visual style and structure vary among chart types, we trained a separate classifier for each chart type. We manually labeled 319 examples (171 bar, 68 line, and 80 pie charts) and split them into train, validation, and test splits with 8:1:1 ratios, respectively.
Our models achieved a precision of 95.0\%  overall and 97.6\% for title classification on our test set. We then used our models to predict the text roles for the remaining charts in the Pew dataset.

We used the extracted title as the final chart title if there was no associated \texttt{alt} text with the chart image. If the \texttt{alt} text was available, we took the longer one by comparing it with the extracted title.

\vspace{0.5em}
\noindent \Nii \textit{Identification of candidate paragraphs:} We observed that relevant paragraphs tend to appear in close proximity to a given chart and share some content with the chart (\eg axis labels, data values). We first used this proximity criteria to form a list of candidate paragraphs $\gL_c$. Specifically, for each chart, we selected the paragraph adjacent to the chart as well as the five paragraphs before and after it as candidates (maximum of 11 in total).

Next, we used a heuristic-based approach to automatically select a subset of relevant paragraphs $\gL_r \subset \gL_c$. 
We
estimated the relevance score of each paragraph in $\gL_c$ to its corresponding chart as $rel = content \times proximity$, where $content$ takes a weighted sum of the number of tokens matched between the paragraph and the OCR-extracted text (numerical tokens were given a higher weight than lexical tokens as they were better indicators of relevance), and $proximity$
\change{is based on the distance between the chart and the paragraph.}
If $rel$ exceeds a threshold and some minimum number of lexical and numerical tokens are matched between the paragraph and chart, we consider such a paragraph to be relevant to the chart.
We set this threshold empirically and chose it to be aggressively high to prioritize precision over recall.
We evaluated the efficacy of our approach against a randomly sampled set of 95 charts and 769 surrounding paragraphs and found a recall of 21.1\% and a precision of 100\%. Given the perfect precision score, we considered the paragraphs in $\gL_r$ to be relevant and to confirm the relevance of the remaining paragraphs, we performed a human study.

\vspace{0.5em}
\noindent \Niii \textit{Selection of relevant paragraphs:} 
We asked crowdworkers on Amazon Mechanical Turk to label how relevant each paragraph is to its chart. 
A total of 5,478 charts and 13,237 paragraphs were annotated. Each chart received two annotations from two 
workers. If both workers labeled a paragraph as either completely irrelevant or relevant (partially/completely), we used the label that they agreed upon as the final label.\footnote{The overall agreement for the crowd workers was 78.2\%.} For  the remaining 2,888 paragraphs where the workers disagreed, we resolved them through internal annotation.

\subsection{Dataset Analysis}

Our chart-to-text datasets contain a diverse range of chart types (\Cref{dataset-chart-types}). Bar charts make up the majority of the charts both in Statista (87.9\%) and Pew (67.9\%) for both simple as well as stacked and group bar charts. The next most common type is line charts
\change{(10.2\% in Statista and 26.4\% in Pew).}

\begin{table}[t]
\centering
\small 
\resizebox{0.75\linewidth}{!}{
\begin{tabular}{lcc|cc}
& \multicolumn{2}{c}{\textbf{Statista}} & \multicolumn{2}{c}{\textbf{Pew}} \\
\midrule
Type & Simple & Complex & Simple & Complex \\
\midrule
Bar &       24,591 &    5,616 &     807 &       5,497 \\
Line &      2,646 &     902 &       325 &       2,129 \\
Area &      0 &         0 &         29 &        105 \\
Scatter &   0 &         0 &         0 &         68 \\
Pie &       409 &       0 &         325 &       0 \\
Table &     223 &       424 &       0 &         0 \\
\midrule
Total &     27,869 &    6,942 &     1,486 &     7,799 \\
\bottomrule
\end{tabular}
}

\caption{Chart type distribution.}

\label{dataset-chart-types}

\end{table}

\begin{table}[t]
\centering
\small 
\resizebox{0.85\linewidth}{!}{
\begin{tabular}{lcc|cc}
& \multicolumn{2}{c}{\textbf{Statista}} & \multicolumn{2}{c}{\textbf{Pew}} \\
\midrule
Statistic & Simple & Complex & Simple & Complex\\
\midrule
\#Vocab. &          39,191 &    18,621 &    9,905   & 18,067 \\
Avg. Character &    295 &       334 &       571     & 635 \\
Avg. Token &        54 &        61 &        110     & 124 \\
Avg. Sentence &     2.56 &      2.62 &      3.84    & 4.27 \\
\bottomrule
\end{tabular}
}

\caption{Chart-to-text dataset statistics.
}
\label{dataset-statistics}
\end{table}

To analyze the topic distribution, we extracted the topic of each chart using its webpage's metadata
(\eg breadcrumbs, meta-tags).
Our datasets cover a broad range of topics including politics, society and health
(see \Cref{topics} in \Cref{app:data-analysis}). The topics in Statista are more evenly distributed than the ones in Pew, which is dominated by \emph{U.S. Politics \& Policy} (45.4\%).

\Cref{dataset-statistics} presents basic linguistic statistics about the datasets. The summaries in Pew are about twice as long as the those in Statista, in terms of average character, token and sentence count. Unsurprisingly, \emph{complex} charts generally have longer summaries than their \emph{simple} counterparts.

\change{We further analyzed the semantic content of the summaries using 100 randomly sampled chart-summary pairs from each dataset.}
\Cref{dataset-content} shows the distribution of sentences across the four  main types of semantic content.\footnote{Our categorization of content is inspired by a recent study \cite{accessibile-viz}.}
We notice that \emph{statistical and comparative} information (\eg min, max, avg.)
is the most common type of content in both datasets. Summaries in Pew tend to report more insights that require more \emph{perceptual and cognitive} efforts (\eg trends and causal relations) which are arguably more challenging to generate compared to simple statistics. Both datasets contain comparable proportions of sentences covering \emph{contextual and domain-specific} information. 
Unlike Statista, Pew summaries rarely explain the chart types and encodings (\eg what do the \texttt{x-} and \texttt{y-} axes represent).

\begin{table}[t]
\centering
\small 
\resizebox{0.9\linewidth}{!}{
\begin{tabular}{lcc}
\textbf{Content Level} & \textbf{Statista} & \textbf{Pew}\\
\midrule
Visual encodings & 32.03\%          &    0.98\%\\
Statistical and comparative &   \textbf{50.00\%} &    \textbf{54.63}\%\\
Perceptual and cognitive &      8.98\%          &    30.49\%\\
Contextual and domain-specific & 10.94\%          &    12.93\%\\
\bottomrule
\end{tabular}
}

\caption{
    \small Distribution of different types of semantic content. 
    }
\label{dataset-content}

\end{table}

We randomly selected 70\%, 15\%, and 15\% of the datasets to create the corresponding train, test and validation splits, respectively.

\section{Chart-to-text Baseline Models}

\paragraph{Problem Definition}

We consider two variations of the chart-to-text problem. In the first variation,  we assume that the underlying data table of the chart is available, where the dataset can be represented as a set of 4-element tuples $\gD = \{\langle C, T, M, S\rangle_n\}_{n=1}^{|\gD|}$ with $C$, $T$, $M$ and $S$ representing the chart image, data table, metadata and textual summary, respectively. For each cell in the data table $T$, we have the following information: \Ni the string value, \Nii the row and column positions, and \Niii whether it is a header cell or not. The metadata $M = (C_{\text{title}}, C_{\text{type}}, C_{\text{labels}})$ consists of the title, type (\eg\ bar, line) and axis labels. 

In the  second variation, we assume that the data table is not available which makes the problem more challenging as well as realistic because most charts online are in image format and do not have the underlying data tables. For a given input $X=\langle C,T,M \rangle$ or $\langle C,M \rangle$, our goal is to generate a textual description $\hat{S}$ which is a good summary of the chart according to a set of evaluation measures.

We consider three categories of models to tackle the task. The first category is image captioning models, where the task is formulated as generating a textual description for the given chart image. The second category is data-to-text models, which rely on the underlying data tables of the charts to produce the corresponding descriptions. Finally, we consider a combination of vision and text models, where the models first extract the 
text 
using the CRAFT OCR model \cite{craftdetection} and then train with a data-to-text setup. 
We present three categories of models below (hyperparameter settings for all the models are provided
in \Cref{app:baselines}).

\subsection{Image Captioning Models} \label{subsec:imgage-cap}
We develop over the Show, Attend, and Tell (SAT) model \cite{Xu2015show} to probe the effectiveness of this category of models for our task. Following \citet{Xu2015show}, we use the ResNet50 \cite{he2016deep} as the image encoder and a unidirectional LSTM \cite{10.1162/neco.1997.9.8.1735} as the decoder for text. 
As the pretrained ResNet50 model is trained on object detection tasks on ImageNet \cite{5206848}, directly applying  it to chart images gave poor results in our experiments. Also, we do not have any object labels for the chart images to train the encoder. Hence, we employ the recently proposed self-supervised strategy called \emph{Barlow Twins} \cite{zbontar2021barlow} which tries to make the embedding vectors of distorted versions of an image sample to be similar, while minimizing the redundancy between the components of these vectors. It achieves state-of-the-art results for ImageNet classification with an accuracy gap of only 3.3\% from the supervised model. We pretrain a separate ResNet50 with Barlow Twins for each of our datasets and use it as an encoder in the 
model.

\begin{figure*}[t!]
\begin{subfigure}[b]{.33\textwidth}
\centering
    \includegraphics[width=1\textwidth,keepaspectratio]{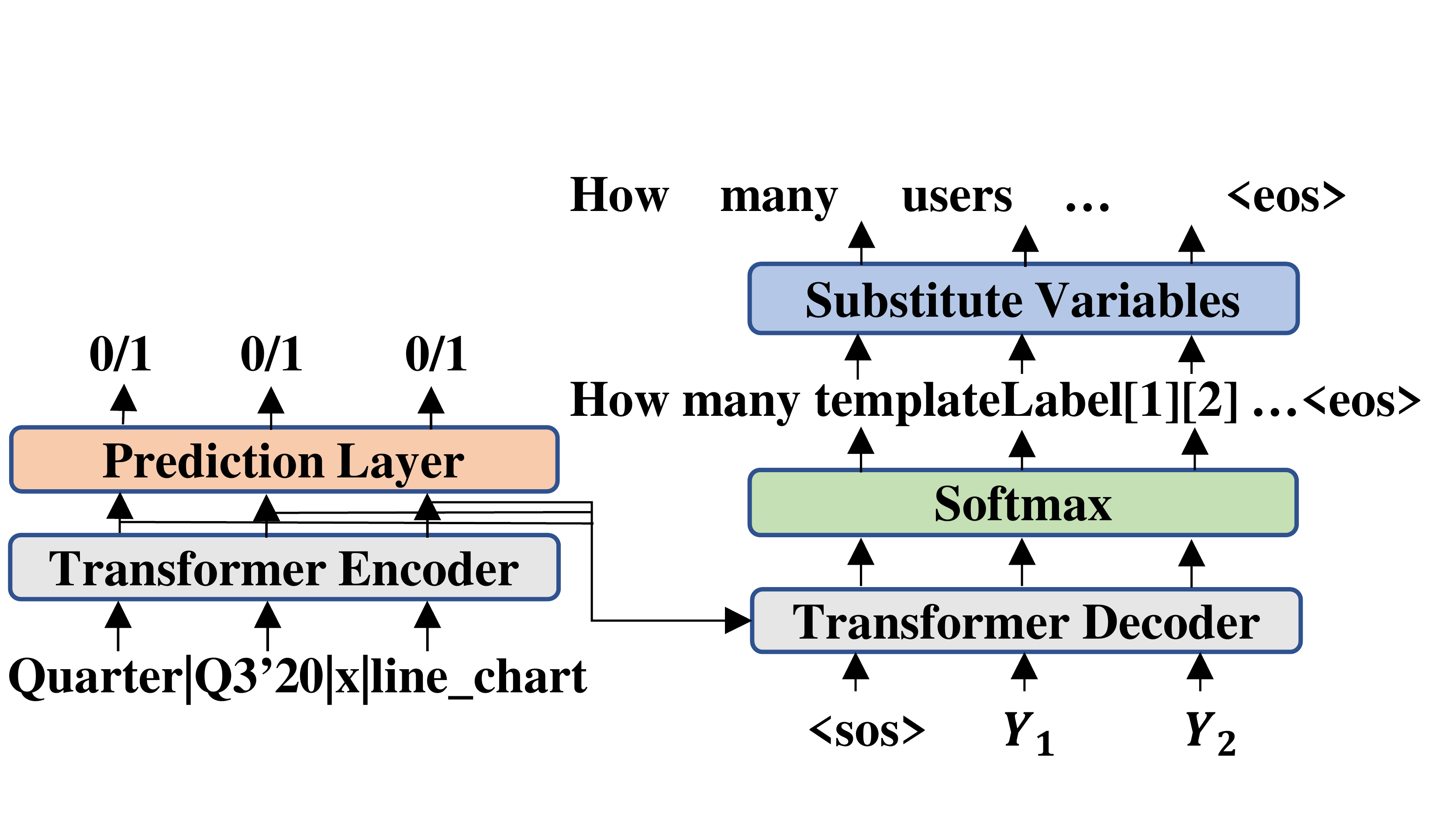}
\caption{\small Chart2text model }
\label{fig:chart2text}
\end{subfigure}
\begin{subfigure}[b]{.33\textwidth}
\centering
    \includegraphics[width=1\textwidth,keepaspectratio]{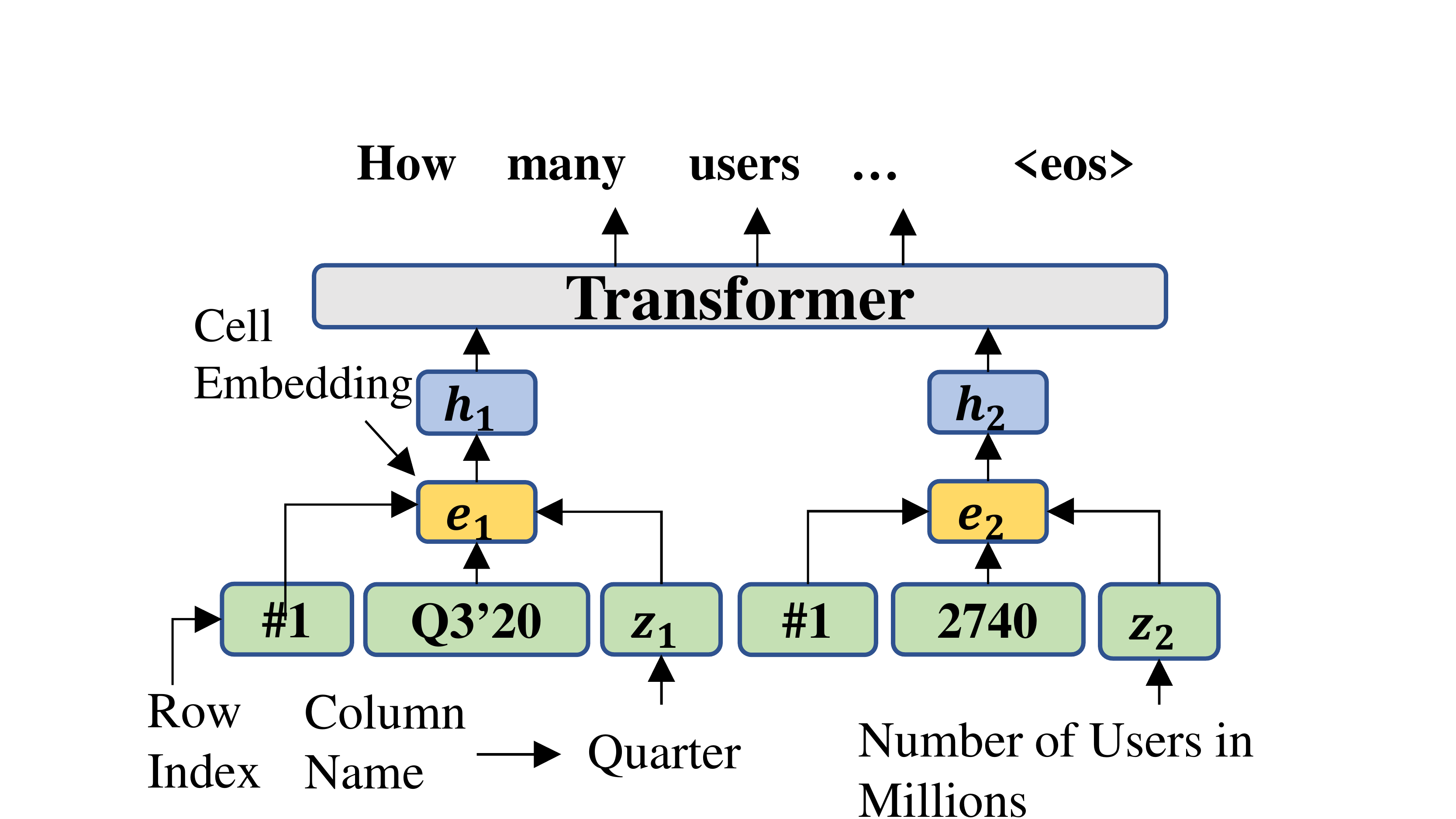}
\caption{\small Field-infusing model}
\label{fig:field-infuse}
\end{subfigure}
\begin{subfigure}[b]{.33\textwidth}
\centering
    \includegraphics[width=0.94\textwidth,keepaspectratio]{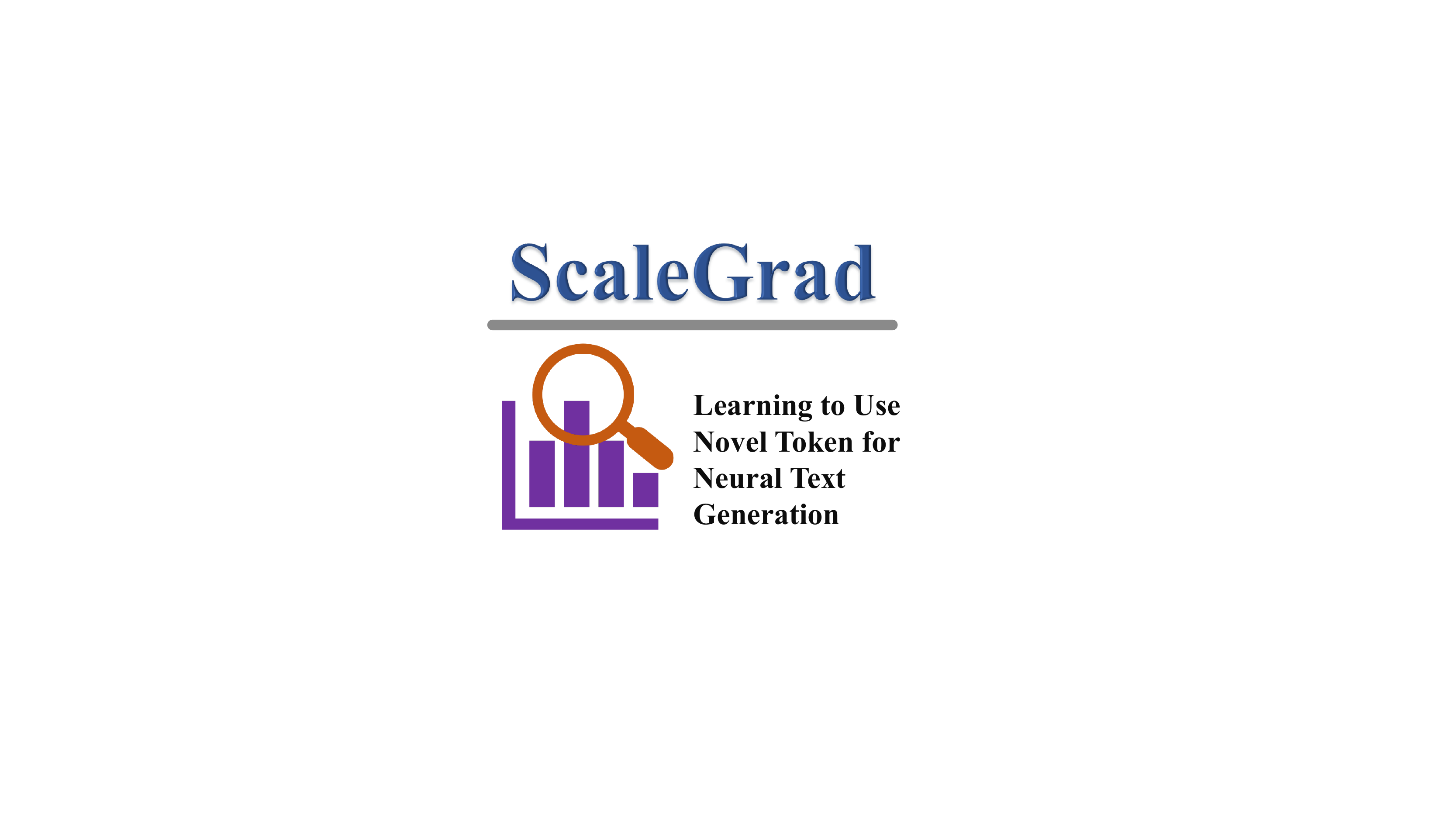}
    
\caption{\small BART/T5 fine-tuning}
\label{fig:bartt5}
\end{subfigure}
\caption{Different chart2text model architectures. \Cref{fig:bartt5} shows fine-tuning stage of the training (not unsupervised pretraining)}
\label{models}
\end{figure*}

\subsection{Data-to-text Models} \label{subsec:data2text}

\vspace{0.5em}
\textbf{$\bullet$ Chart2text}
\cite{obeid} 
is an adapted transformer model for chart-to-text based on the data-to-text model of  \citet{gong-etal-2019-enhanced}. It takes a sequence of data records as input with each record being a set of tuples (\eg\ column header, cell value, column index) and embeds them into feature vectors with positional encodings to distinguish orders (\Cref{fig:chart2text}). The model includes an auxiliary training objective (binary labels indicating the presence of the record in the output sequence) on the encoder to maximize the content selection score. It also implements a  templating strategy of target text with data variables (\eg\
\emph{cells}, \emph{axis labels}) to alleviate hallucination problems. Since in Pew data tables are not available,
we use OCR-generated texts as inputs  which are linearized and embedded into feature vectors. 
The bounding box information of OCR-generated data of each chart is also embedded and concatenated to the table vectors to provide positional information to the model.

\vspace{0.5em}
\noindent \textbf{$\bullet$ Field-Infusing Model}
\cite{chen2020logical}
is inspired by the concept-to-text work \cite{lebret-etal-2016-neural}. The values in a cell are first encoded with an LSTM, which is then concatenated with the embeddings of row index and column heading. These table representations ($h_1, h_2$ in \Cref{fig:field-infuse}) are then fed into a 3-layer Transformer encoder-decoder model to generate the target summaries. Additionally, for Pew, 
we embed the bounding box information of the chart OCR-texts and concatenate it to the LSTM-based field representation as an auxiliary positional information to the model.

\vspace{0.5em}
\noindent  \textbf{$\bullet$ BART}
\cite{lewis-etal-2020-bart} 
adopts a seq2seq Transformer architecture with  denoising pretraining  objectives. It is particularly pretrained to be effective  for text generation tasks. 
For our chart-to-text tasks, 
we flatten the data table row by row and concatenate  the title with table content as the input to the encoder (\Cref{fig:bartt5}). {In the absence of data tables, we concatenate all the OCR-texts in a top to bottom order and fed it to the model as input.}

\vspace{0.5em}
\noindent  \textbf{$\bullet$ T5}
\cite{Raffel2020t5} is a unified seq2seq Transformer 
model 
that converts various NLP tasks into a text2text generation format. It is first pretrained with a `fill-in-the-blank' denoising objective, where 15\% of the input tokens are randomly dropped out. The spans of consecutive dropped-out tokens are replaced by a sentinel token. The decoder then has to predict all of the dropped-out token spans, delimited by the same sentinel tokens used in the input. This is different from the pretraining objective of BART where the decoder predicts the entire original sequence (not just the dropped spans). T5 is  fine-tuned with several supervised multi-task training objectives (\eg machine translation, text summarization). We format the input in the same way as for the BART models. Specifically, we add ``translate Chart to Text: " to the prefix of the input to mimic the pretraining process (see \Cref{fig:bartt5}).

{For OCR-based input, we experiment with two T5 model variants. In the first variant, we concatenate all the OCR-extracted sentences from the chart image in a top to bottom order and fed it to the model as input. In the second, we modify the input to accommodate the spatial information of the detected texts. Inspired by \citet{lxmert}, we feed the bounding box coordinates of each detected text token into a linear layer to produce  positional embeddings which are then added to their corresponding embeddings of the OCR tokens as input.}

\section{Evaluation}

\subsection{Automatic Evaluation}

\paragraph{Measures} For automatic evaluation of the summary quality, we utilized five measures. BLEU \cite{post-2018-call} and CIDEr \cite{vedantam2015cider} measure n-gram overlaps between the model generated text and the reference text. CIDEr computes TF-IDF weighted n-gram overlaps. BLEURT \cite{sellam2020bleurt} is a model-based evaluation metric that indicates to what extent the candidate is grammatical and conveys the meaning of the reference. We use BLEURT-base-128.    
Content Selection (CS) metric   measures how well the generated summaries match the gold summaries in terms of selecting records to generate~\cite{wiseman2017challenges}. Since both the BLEURT and CS are calculated at the sentence-level, we 
average these scores over the whole test set.
Finally, for 
readability and fluency, we measure Perplexity (PPL) using a pre-trained  GPT-2 Medium \cite{radford2019language}.

\begin{table}[t!]
    \centering
    \scalebox{0.56}{\begin{tabular}{lccccc}
        \textbf{Models} & \textbf{BLEU} $\uparrow{}$ & \textbf{CS} $\uparrow{}$ & \textbf{BLEURT} $\uparrow{}$ & \textbf{CIDEr} $\uparrow{}$ & \textbf{PPL} $\downarrow{}$ \\
        \midrule
        \multicolumn{6}{c}{\textbf{Statista}}  \\
        {Image Caption} & 15.94 & 25.70\% & -0.76 & 0.95 & 10.53 \\
        {TAB-Chart2text} & 21.10 & 56.10\% & 0.06 & 2.61 & 28.79 \\
        {TAB-Field-Infuse} & 12.09 & 42.07\% & -0.32 & 1.78 & 17.01 \\
        {TAB-BART} & 36.36  & \bf{77.14\%} & 0.12 & 4.40 & 12.55 \\
        {TAB-T5} & \bf{37.01}  & 75.72\% & \bf{0.15} & \bf{4.68} & 10.00 \\
        {OCR-T5} & 35.29 & 73.77\% & 0.10 & 4.43 & 8.66 \\
        {OCR-T5}$^{\star}$ & 34.55 & 73.55\% & 0.09 & 4.37 & \bf{8.59} \\
        \change{TAB\_OCR-Chart2text} & 7.64 & 47.58\% & -0.44 & 1.09 & 54.98 \\
        \change{TAB\_OCR-Field-Infuse} & 7.03 & 37.63\% & -0.49 & 1.18 & 14.76 \\
        \change{TAB\_OCR-BART} & 35.83 & 72.15\% & 0.09 & 3.97 & 13.99\\
        \change{TAB\_OCR-T5} & 36.74 & 72.22\% & 0.13 & 4.33 & 10.20\\
        \midrule
        \multicolumn{6}{c}{\textbf{Pew}} \\
        {Image Caption} & 4.09 & 2.14\% & -0.96 & 0.38 & 16.43 \\
        {OCR-Chart2Text}$^{\star}$ & 7.20 & 24.49\% & -0.56 & 0.65 & 12.11 \\
        {OCR-Field-Infuse}$^{\star}$ & 0.19 & 10.12\% & -1.01 & 0.26 & 9.57 \\
        OCR-BART &  9.09 & 39.99\% &  -0.38 &  1.97 & 11.04 \\
        {OCR-T5} & \bf{10.49} & \bf{40.87\%} &  \bf{-0.35} &  \bf{2.20} & 10.11 \\
        {OCR-T5}$^{\star}$ & 10.42 & 40.31\% & -0.42 & 2.13  & \bf{8.65} \\
        \bottomrule
    \end{tabular}}
    \caption{Evaluation results for different models on Statista and Pew test sets. $\uparrow{}$: Higher is better, $\downarrow{}$: Lower is better. {``TAB- " models have access to the underlying data table and ``OCR- " models use OCR-extracted data. OCR variants with $\star$ superscript use bounding box information. \change{``TAB\_OCR- " models use automatically generated data tables.}}
    }
\label{tab:evaluation-table}
\end{table}

\paragraph{Results}

In general, from the results in Table~\ref{tab:evaluation-table}, we notice that large-scale unsupervised pretraining (\ie ``~-BART", ``~-T5") helps to boost the performance significantly. In terms of the model variants, the image captioning model has failed to capture relevant information from charts (low CS score) even though it generates fluent text (low PPL).

On Statista, when the data tables are available,  Chart2text and Field-Infuse models are able to extract information from the data table, but they struggle to produce texts with good quality. {This could be because these models did not use any large-scale pretraining.} On the other hand, TAB-BART and TAB-T5 are able to produce well-structured and relevant summaries. The OCR-based models can generally generate fluent summaries but they are slightly less effective in extracting the relevant information since the OCR process introduces some noise in the input data. 

\change{ We also experiment with automatically extracted tables to see how the models perform in the absence of gold data tables. To this end, we extended ChartOCR \cite{ChartOCR}, which predicts the raw data values of chart elements, to extract the fully-structured data table. The accuracy of automatic data extraction was 77.31\% (see Appendix \ref{app:data_extraction} for details). We find that similar to OCR-based models, TAB\_OCR-based models tend to be less effective in extracting the relevant information compared to their TAB-based counterparts which use ground truth data tables. 
}

Pew, on the other hand, is much challenging because it contains many charts with ill-defined structure and the underlying data tables are not available. Unsurprisingly, the performance of all the models has dropped significantly compared to that on Statista. Nonetheless, we can see that without the presence of the underlying data table, the vision+text (OCR-based) models have brought notable improvements over the vision only model. {Further breakdown of model performance based on chart types is provided in \Cref{app:charttype}.}

We also evaluate the \emph{transferability} of the models and the datasets, where we first pretrain a model on a source dataset and fine-tune it on the target dataset. In addition to our two datasets (Statista or Pew), we experiment with ToTTo \cite{parikh2020totto} as another source dataset, which is a large-scale open-domain English table-to-text dataset.
Our results show that pretraining on other datasets only brings about marginal improvement. Details of this experiment can be found in \Cref{app:transfer}.

\begin{table*}[t!]
    \centering
    \small
    \scalebox{0.9}{\begin{tabular}{lccc|ccc|ccc}
    \toprule
    & \multicolumn{3}{c}{\textbf{TAB-T5 (1) vs. OCR-T5 (2)}} & \multicolumn{3}{c}{\change{\textbf{Gold (1) vs. TAB-T5 (2)}}} & \multicolumn{3}{c}{\change{\textbf{Gold (1) vs. OCR-T5 (2)}}} \\ 
    \cmidrule(r){2-4} \cmidrule(l){5-7} \cmidrule(l){8-10}
    \textbf{Summary} & \textbf{Factual} & \textbf{Coherence} & \textbf{Fluency} & \textbf{Factual} & \textbf{Coherence} & \textbf{Fluency} & \textbf{Factual} & \textbf{Coherence} & \textbf{Fluency} \\
    \midrule
    Summary 1 Win &    55.3\% &    23.3\% &    20.0\% &    30.0\% &    36.7\% &    22.0\% &    59.3\% &    43.3\% &    28.7\% \\
    Summary 2 Win &    12.0\% &    11.3\% &    11.3\% &    13.3\% &    16.7\% &    14.0\% &    7.33\% &    15.3\% &    17.3\% \\
    Tie &           32.7\% &    65.3\% &    68.7\% &    56.7\% &    46.7\% &    64.0\% &    33.3\% &    41.3\% &    54.0\% \\
    \midrule
    $p$-value (sign test) & 1.86e-11 & 8.77e-3 & 0.0395 & 1.31e-3 & 5.26e-4 & 0.0668 & 1.27e-16 & 4.25e-6 & 0.0266 \\
    \bottomrule
    \end{tabular}}
    \caption{Human evaluation results for comparing between the outputs of TAB-T5, OCR-T5 and the gold summary.}
    \label{tab:evaluation-table-human}
\end{table*}

\begin{figure*}[t] 
\linespread{0.5}\selectfont\centering 
\scalebox{.99}{\begin{tabular}{p{3.5cm} | p{3.7cm} | p{3.5cm} | p{3.5cm}}     
\toprule
        \raisebox{-.9\height}{\includegraphics[height=2cm]{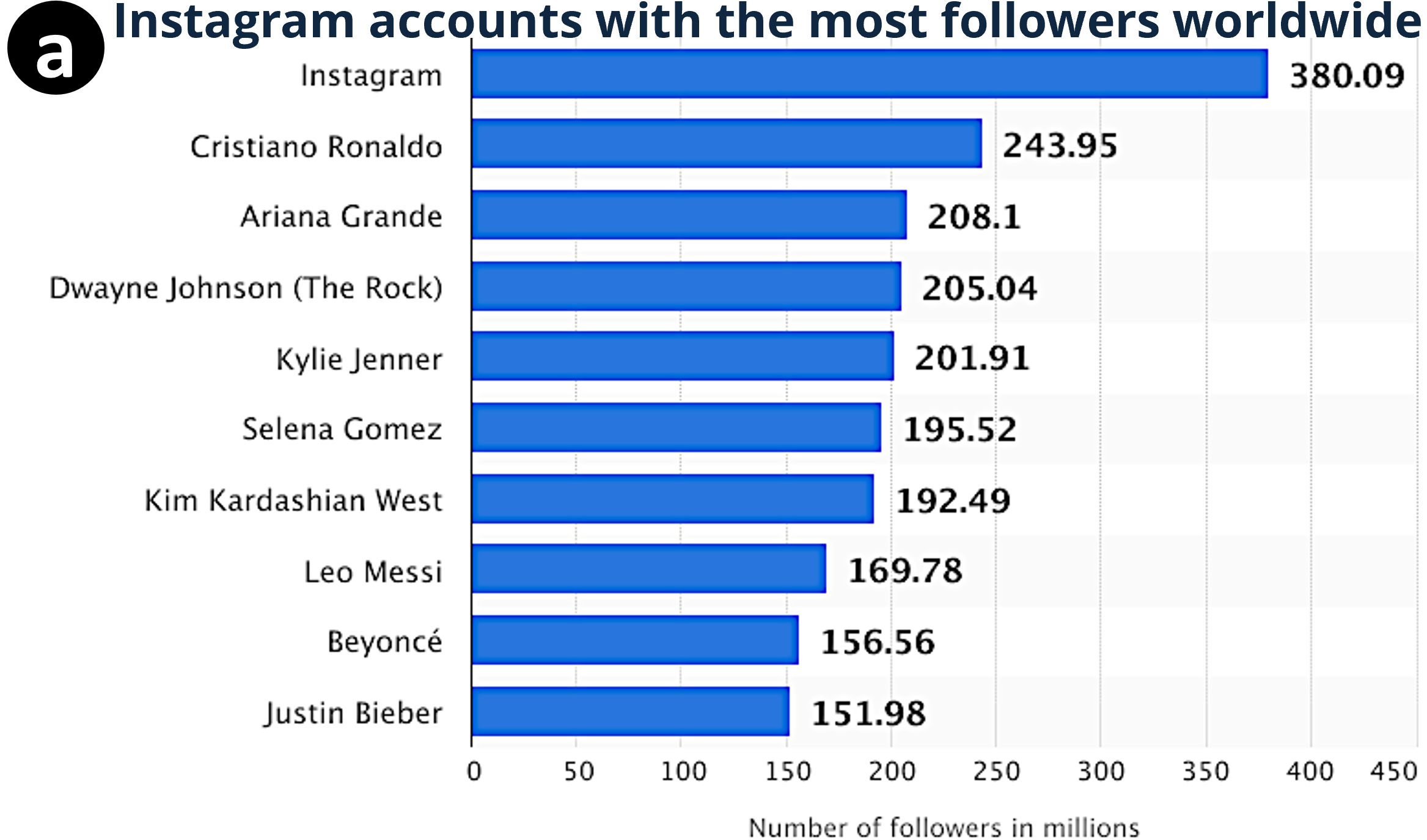}}
        &
        {\tiny \textbf{Gold:} Footballer Cristiano Ronaldo heads the ranking of the most popular Instagram accounts as of December 2020. He is the most-followed person on the photo sharing app platform with nearly 244 million followers. Instagram's own account was ranked first with over 369 million followers.}
        &
        {\tiny \textbf{TAB-T5} As of December 2020, the Instagram account with the most followers worldwide was Cristiano Ronaldo's account with 243.95 million followers. The social network's most popular account was Instagram, which had 380.09 million followers.}
        &
        {\tiny \textbf{OCR-T5:} As of December 2020, Cristiano Ronaldo was the Instagram account with the most followers worldwide, with \textcolor{blue}{380.09} million followers. The \textcolor{red}{Brazilian singer had 243.95 million followers on the photo sharing app.}}
        \\
\midrule 
        \raisebox{-.9\height}{\includegraphics[height=2.4cm]{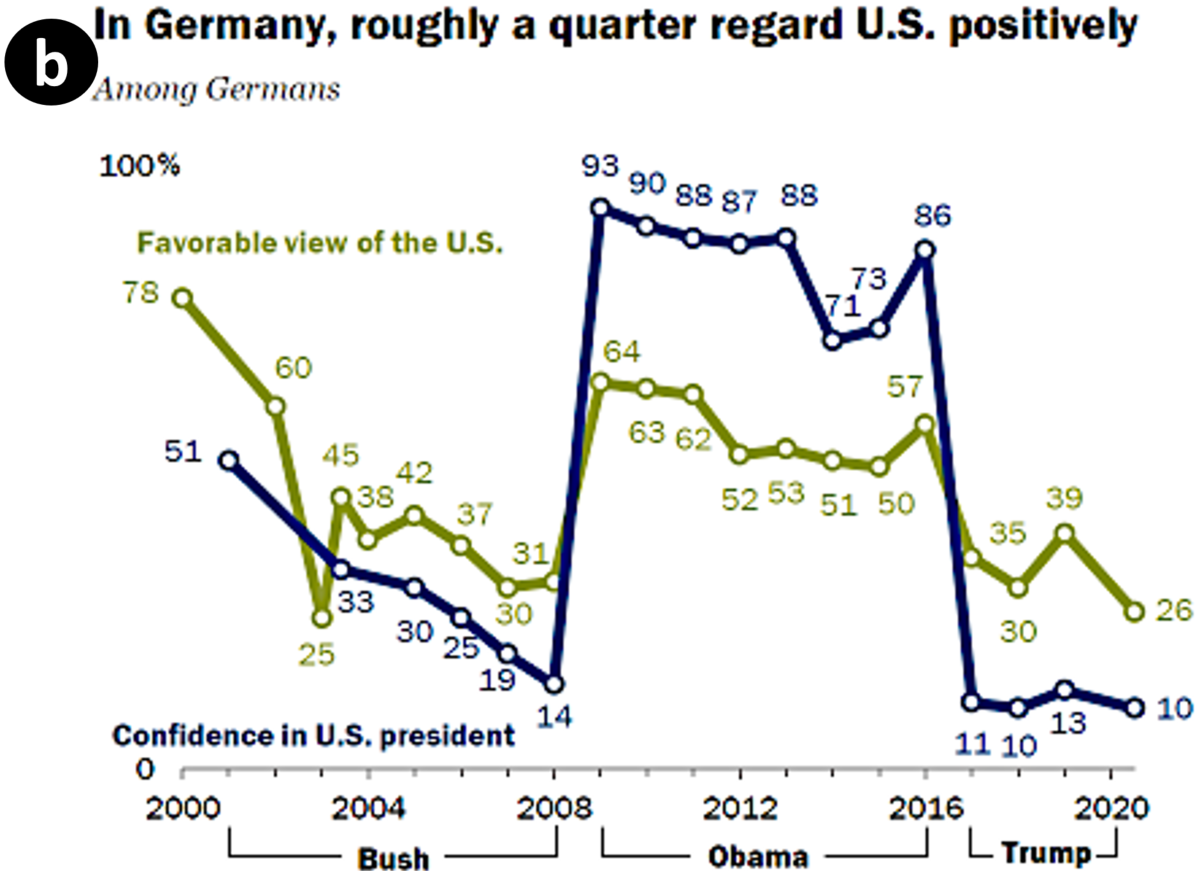}}
        & 
        {\tiny \textbf{Gold:} Germans give the U.S. some of its worst ratings in the survey. Only 26\% have a positive view of America, while just 10\% have confidence in Trump when it comes to his handling of world affairs. These opinions are in stark contrast to the very favorable assessments Germans had during Barack Obama’s presidency, but roughly on par with views at the end of George W. Bush’s tenure.}
        &
        {\tiny \textbf{OCR-BART} \textcolor{red}{Roughly nine-in-ten Germans (88\%) express a favorable view of the U.S. This is up sharply from 73\% in 2016}, but it also is the highest percentage since Pew Research Center began surveying in Germany in 2000. Germans’ views of America have also become more positive since the end of the George W. Bush era..}
        &
        {\tiny \textbf{OCR-T5} Germany is a country where U.S. image has declined sharply in recent years. \textcolor{red}{For example, in 2008, before the coronavirus outbreak began to spread, about a third of Germans had confidence in President Barack Obama to do the right thing in world affairs. By 2014, confidence in Obama had fallen to about a third, but by 2019 confidence had nearly doubled to about a quarter.}}
        \\
\midrule  
        
        \raisebox{-.9\height}{\includegraphics[height=2.4cm]{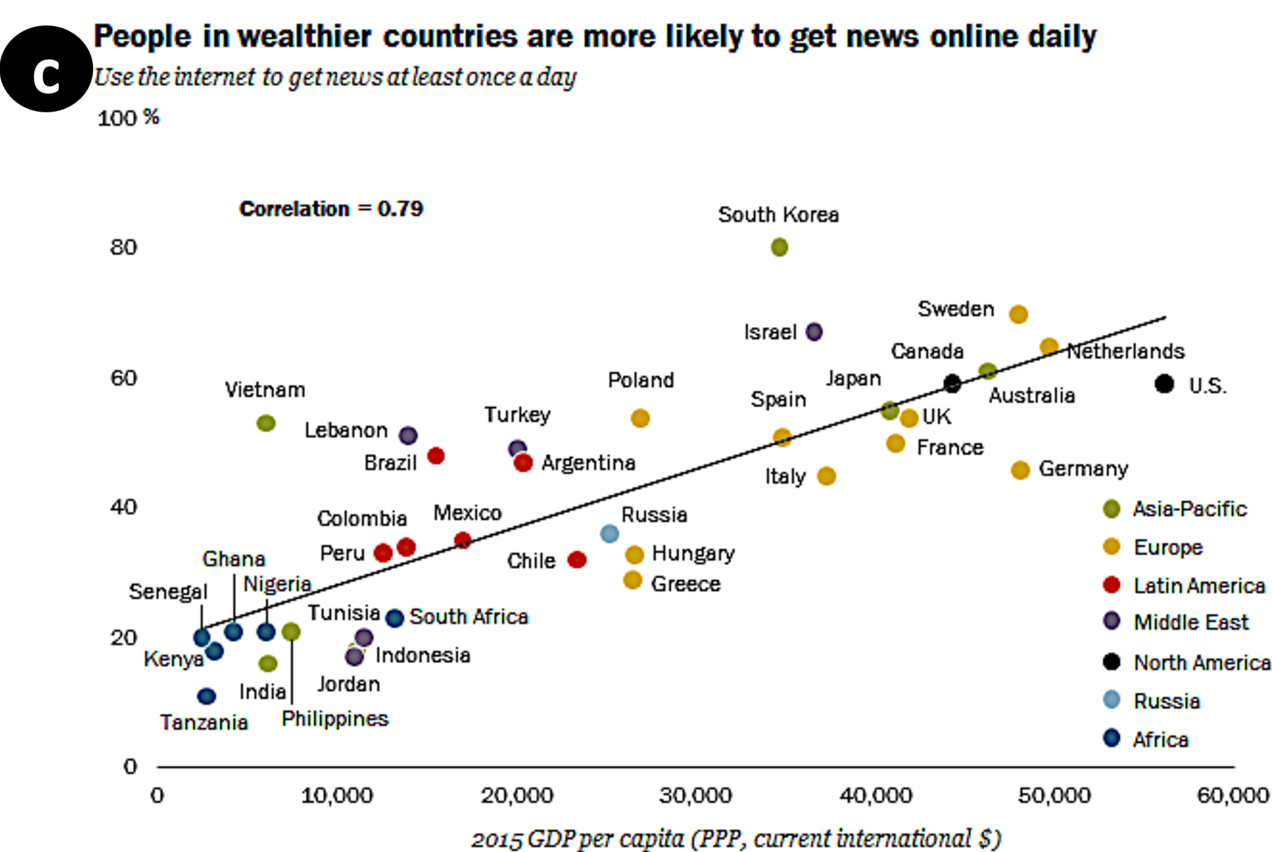}}
        &
        {\tiny \textbf{Gold:} In general, internet access has been shown to be higher in wealthier countries, and this plays out to a greater likelihood of using the internet for news as well. For example, 61\% in Australia – which had a 2015 gross domestic product (GDP) per capita of \$46,271 – get news at least once a day through the internet. Just 20\% in Senegal, with a GDP per capita of \$2,421, do the same.}
        &
        {\tiny \textbf{OCR-BART} People in wealthier countries are more likely than those in poorer nations to get news online at least once a day. This is especially the case in \textcolor{blue}{Venezuela, where about seven-in-ten people (71\%)} get news daily.}
        &
        {\tiny \textbf{OCR-T5} However, it is not the case that the correlation between digital news consumption and personal income is not quite so strong. The study found that in wealthier countries, people were more likely to use the internet for news on a daily basis than in poorer countries. The finding that people in wealthier countries tended to do this more often than those in poorer nations to get news online.}
        \\
\bottomrule  
    \end{tabular}
    }
    \caption{Sample outputs from Statista (first column) and Pew datasets (last two columns). \red{Red} indicates hallucination errors and \blue{blue} indicates  tokens that are resulted in factual errors in the model output.
    }
    \label{tab:examples}
\end{figure*}

\subsection{Human Evaluation}

To further assess the summary quality 
we performed a human evaluation on 150 randomly sampled charts from the Statista dataset with four internal annotators who are native speakers of English.  For each chart, annotators performed pairwise comparisons between the outputs of TAB-T5, OCR-T5 and the original gold summary (served as a control), resulting in a total of 450 pairwise comparisons (\Cref{app:human-eval}). They compared the summaries based on three criteria: \Ni \textbf{Factual correctness}: Which summary is more factually correct (\ie\ facts mentioned are supported by the chart)? \Nii \textbf{Coherence}: Which summary is more coherent (\ie\ sentences are well connected)? and \Niii \textbf{Fluency}: Which summary is more fluent and grammatically correct? For each criterion, the annotator 
picked the better one (win) or equally good (tie). {Each comparison was performed by one annotator, except the first 150 comparisons for which we had two annotators to measure the agreement. The  agreement for these 150 comparisons, excluding ties, was 74.3\% (ties were excluded since they do not affect the overall ranking of the summaries).}

Table~\ref{tab:evaluation-table-human} shows that the TAB-T5 performed significantly better than OCR-T5 based on all three criteria, especially on factual correctness. This is likely because, without the data table as input, OCR-T5 model often fails to generate factually correct statements from the OCR text. \change{We also observe that while the fluency of the model outputs is comparable to the gold summary, their factual correctness and coherence were significantly worse, especially for the OCR-T5 model}.

\subsection{Error Analysis and Challenges}
\label{sec:error-analysis}

We manually analyzed 200 random samples from Statista and Pew. We chose TAB-T5 and OCR-T5 for Statista and OCR-BART and OCR-T5 models for Pew. This analysis helps us to understand model errors and identify key challenges that existing models face as we describe below.

\vspace{0.5em}
\noindent \textbf{Perceptual and reasoning aspects~} \change{As mentioned in \cref{sec:intro},} charts often describe complex patterns and trends which can be perceived by humans easily but they are not necessarily easy to derive through \change{analysis of raw data tables}. 
In ~\Cref{tab:examples}b, the OCR-T5 model manages to describe a trend correctly in the first sentence but describes a trend incorrectly in the last sentence. \change{These examples demonstrate the shortcomings of existing models.} \change{In order to explain perceptual and reasoning aspects effectively, we need  more sophisticated models that better capture prominent visual relationships in charts. In particular, we aim to develop better representations 
including semantic graph representation of the chart that encodes numerical and logical relationships among chart objects}.

\vspace{0.5em}
\noindent \textbf{Hallucinations~} Sometimes, the model outputs tokens that are irrelevant to the chart.
For example, while the model outputs 
in \Cref{tab:examples}a,b are quite fluent, they contain hallucination errors. This problem is commonly observed in other  data-to-text work as well \cite{wiseman2017challenges,parikh2020totto}.

\vspace{0.5em}
\noindent \textbf{Factual errors~} 
Factually incorrect statements are more common for the OCR-based models (\eg\ 
in ~\Cref{tab:examples}a-b) since they do not take the data table as input, thus fail to associate the data values correctly. In contrast, TAB-T5 which utilizes the data table as input tends to 
generate less factual errors.
This confirms that summarizing charts when the data table is not available is usually more challenging.

\vspace{0.5em}
\noindent \textbf{Computer vision challenges~} The factual errors  illustrate some unique computer vision challenges. First, charts do not always show data values as text labels, 
thus the OCR models cannot access those 
values. Even if the  data values are labeled, the absence of association between data values (\eg\ \texttt{Instagram} is related to \texttt{380.09M} 
in ~\Cref{tab:examples}a) leads to factual errors. 
This problem might be alleviated if the model can extract the data table from a chart image. While there are some initial attempts in this direction (\eg\ \citet{ChartOCR,  Choi2019VisualizingFT}), 
more accurate data extraction from charts is necessary. 

\vspace{0.5em}
{\noindent \textbf{Generalizability~} The charts in our benchmark cover several different chart types and a wide variety of topics (\cref{topics}). The charts in the Pew in particular have a wide variety of visual styles in terms of color, layout and typography as they were created over several years by different authors (see examples in \cref{tab:example}). Nevertheless,  finding more chart-summary pairs with more diverse visual styles is  an open challenge. In future, we aim to find more different sources of chart-summaries and perform cross-domain experiments across those different sources to evaluate the generalizability of models.}

\section{Conclusion}

We have presented two large-scale datasets for chart summarization. We also provided several state-of-the-art baselines and measures. Our evaluation highlights the promise of these baselines and also reveals several unique challenges for the chart summarization task. We hope that Chart-to-text will serve as a useful research benchmark for model and metric development and motivate other researchers to explore this relatively new area. 
\section*{Acknowledgement}
The authors would like to thank the anonymous reviewers for their helpful comments. This research was supported by the Natural Sciences \& Engineering Research Council (NSERC) of Canada.
\section*{Ethical Considerations}
During the dataset collection and annotation process, we had many ethical issues to take into consideration. To respect the intellectual property of the chart publishers, we only used publicly available charts from resources that provide publication rights of downloaded content for academic purposes. According to the terms of use and publication rights for Statista,\footnote{\href{https://www.statista.com/getting-started/publishing-statista-content-terms-of-use-and-publication-rights}{https://www.statista.com/getting-started/publishing-statista-content-terms-of-use-and-publication-rights}} users are granted publication rights only to free studies of Statista, so we only used the free publicly available webpages. According to the terms and conditions for Pew,\footnote{\href{https://www.pewresearch.org/about/terms-and-conditions/}{https://www.pewresearch.org/about/terms-and-conditions/}} users are allowed to use the content as long as they are attributed to the Center or are not attributed to a different party. 

To fairly compensate the Mechanical Turk annotators, we compensated the annotators based on the minimum wage in the United States at the time (7.25 US\$ per hour) and the estimated time taken for each task (1 minute). Hence, these annotators received 0.10 - 0.15 US\$ for each chart, depending on the number of candidate paragraphs associated with it. Additionally, to protect the privacy of these annotators, all of their annotations were anonymized.

To ensure the reproducibility of our experimental results, we have provided the hyperparameter settings and estimated training time in \Cref{app:baselines}.

We foresee one possible misuse of our models that is to spread misinformation. Currently, our model outputs tend to appear fluent but contain some hallucinations and factual errors, as detailed in \Cref{sec:error-analysis}. Hence, if such model outputs are published without being corrected, it may mislead and misinform the general public.



\bibliographystyle{acl_natbib}
\bibliography{chart2text}
\newpage
\appendix
\section{Appendices}
\subsection{Additional Details on Data Annotation}
\label{app:annotation-details}
\subsubsection{Example Webpage from Statista}
\label{app:statista-first-caption}
\begin{figure*}
    \centering
    \includegraphics[width=0.9\textwidth]{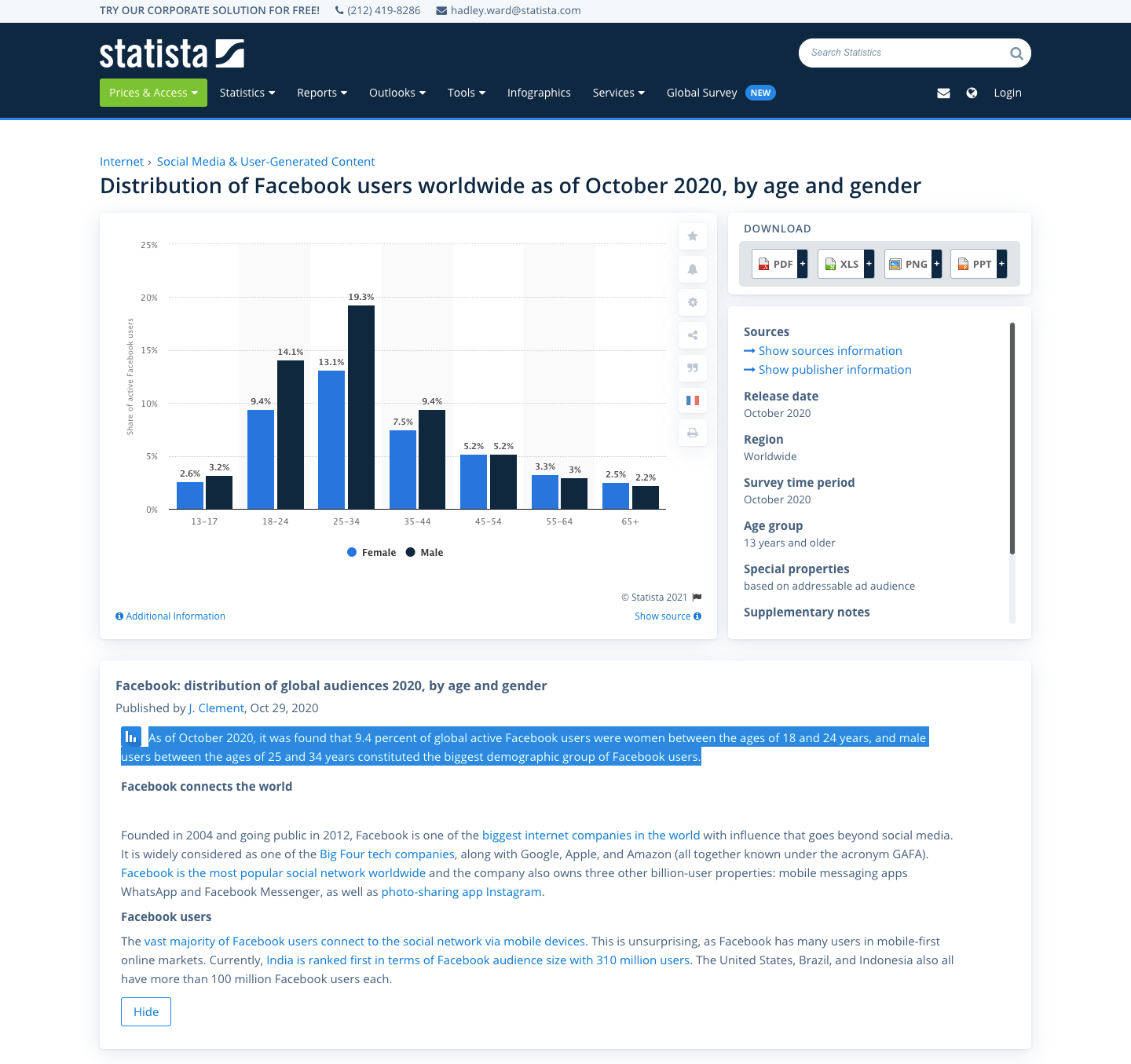}
    \caption{A screenshot of a webpage from Statista.}
    \label{statista-first-caption}
\end{figure*}

An example of a webpage from Statista is given in \Cref{statista-first-caption}. It contains a chart image and its accompanying description text. The first part of the text (highlighted in blue) provides a succinct summary of the chart while the remaining parts of the text (not highlighted) provides irrelevant background information, such as Facebook's history.

\subsubsection{Annotation of x-axis Labels in Statista}
\label{app:statista-annot-interface}
\begin{figure*}
    \centering
    \includegraphics[width=0.9\textwidth]{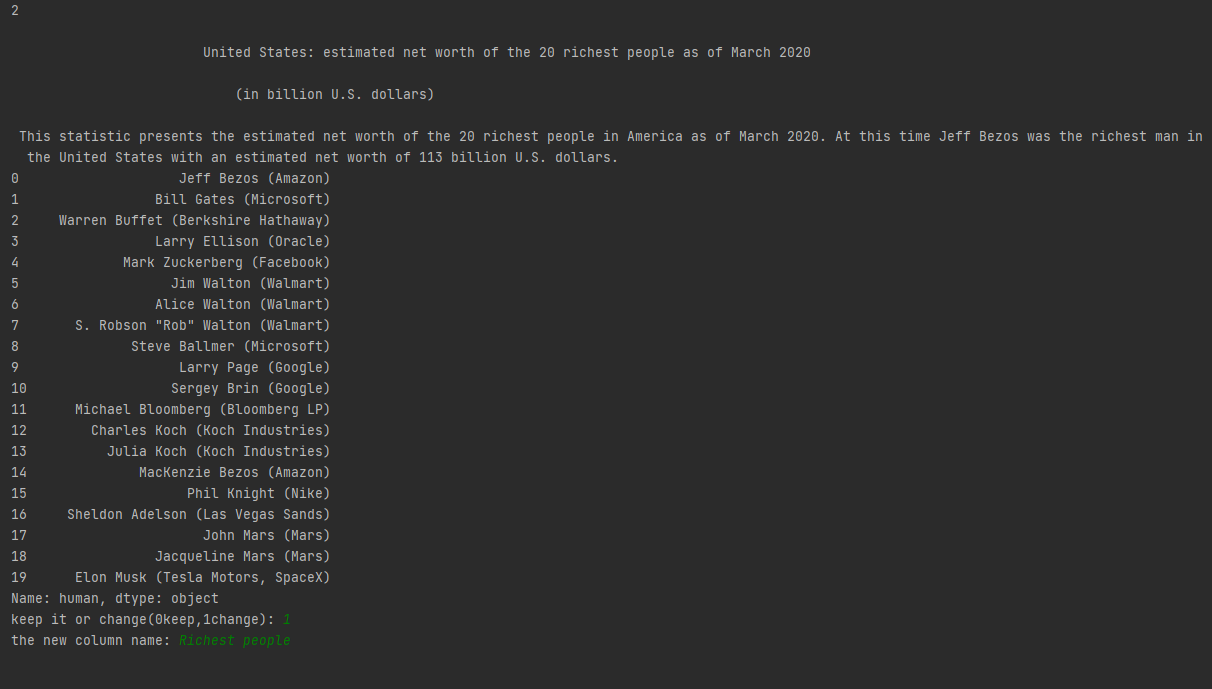}
    \caption{The user interface for labeling the x-axis labels in the Statista dataset.}
    \label{statista-annot-interface}
\end{figure*}

The user interface for the annotation task of labeling the x-axis labels in the Statista dataset is given in \Cref{statista-annot-interface}.

\subsubsection{Identify Candidate Paragraphs in Pew}
\label{app:pew-relevance-score}
The details for computing the relevance score of a paragraph to the given chart, and the heuristic for finding relevant paragraphs in the Pew dataset are given in \Cref{pew-relevance-score}.

\begin{figure*}
\centering
\begin{equation*}
    \begin{split}
        & \text{Let }s_i\text{ be the relevance score for sentence }i\text{ in the paragraph.}\\
        & \text{Let }l_i\text{ be the number of lexical token matches between sentence }i\text{ and the chart.}\\
        & \text{Let }n_i\text{ be the number of numerical token matches, excluding year tokens, between sentence }i\text{ and} \\ &\text{the chart.}\\
        & \text{Let }y_i\text{ be the number of year token matches between sentence }i\text{ and the chart.}\\
        & \text{Let }u_i\text{ be the number of numerical tokens that appear in sentence }i\text{ but not in the chart.}\\
        & \text{Let }c\text{ be the number of sentences in the paragraph.}\\
        & s_i = 0.58l_i + 1.4n_i - 0.5u_i \\
        \\
        & \text{Let }content\text{ be the content score of the paragraph.}\\
        & content = \frac{1}{1 + \exp{\left(0.3 \times \left(-\underset{i}{\max}\left(s_i\right) + 1.7\right)\right)}} \\
        \\
        & \text{Let }proximity\text{ be the proximity score of the paragraph.}\\
        & \text{Let }dist\text{ be the proximity of the paragraph to the chart.} -5 \leq dist \leq 5\\
        & \text{For example, }dist = -1\text{ if the paragraph is directly before the chart, }dist = 0\text{ if it contains the chart }\\
        & \text{and}~ dist = 1 \text{ if it is directly after the chart.}\\
        & proximity = 0.4 \times \exp{\left(-0.1 \times {\lvert dist \rvert}^2\right)} + 0.6\\
        \\
        & \text{Let }rel\text{ be the relevance score of the paragraph.}\\
        & rel = content \times proximity\\
        \\
        & \text{Heuristic: A paragraph is relevant if it satisfies the following conditions:}\\
        & \sum_i l_i > 3\\
        & \sum_i n_i + y_i > 0\\
        & \sum_i u_i = 0\\
        & rel > 0.72\\
        & c > 0\\
    \end{split}
\end{equation*}
\caption{The computation of a paragraph's relevance score to a chart, and the conditions for the heuristic in the Pew dataset.}
\label{pew-relevance-score}
\end{figure*}

\subsubsection{Relevant Paragraph Selection in Pew}
\label{app:pew-annot-interface}
For the relevant paragraph selection task, the annotators received 0.10 - 0.15 US\$ for each chart, depending on the number of candidate paragraphs associated with it. To ensure the quality, we recruited participants with at least 95\% approval rate and 5000 approved HITs (Human Intelligence Tasks) and they were only allowed to complete the tasks after they successfully completed a sample task.

\begin{figure*}
    \centering
    \includegraphics[width=0.8\textwidth]{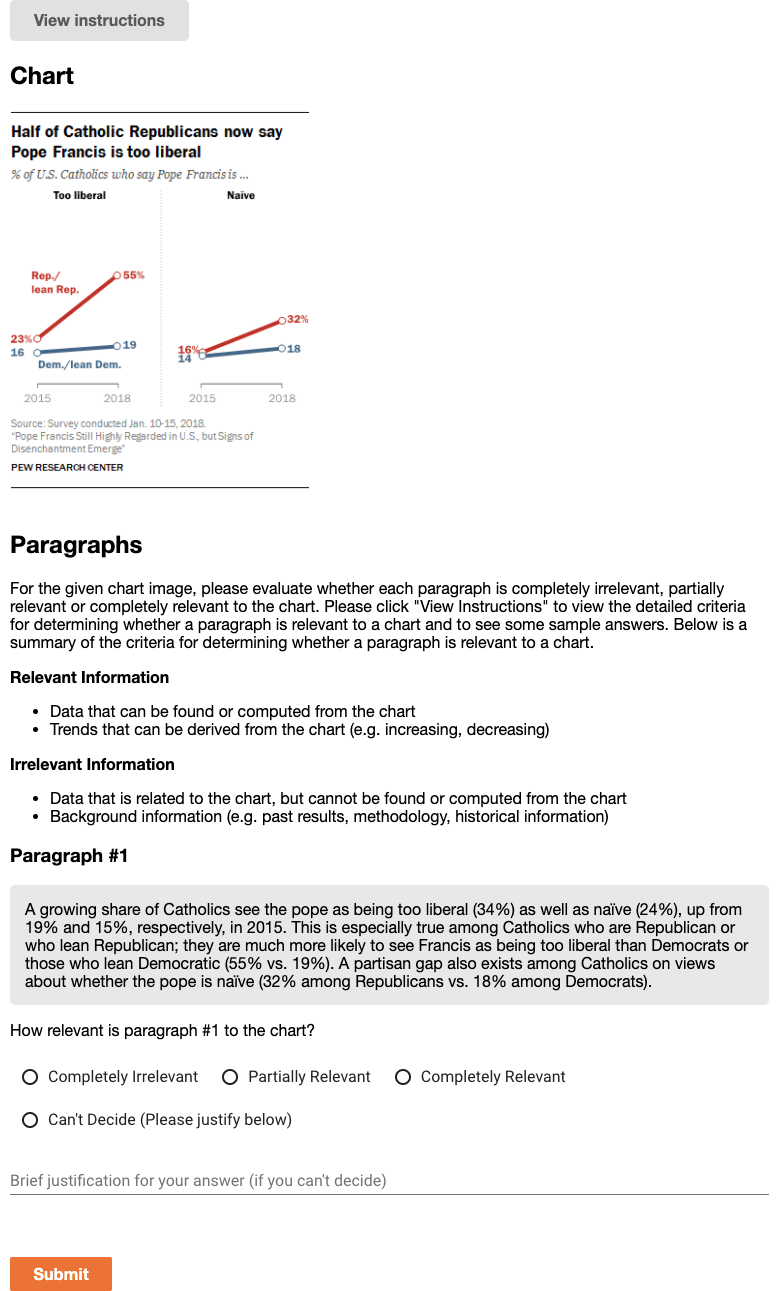}
    \caption{The user interface for the Mechanical Turk annotation task in the Pew dataset.}
    \label{pew-annot-interface}
\end{figure*}

The user interface for the Mechanical Turk annotation task of selecting paragraphs relevant to charts in the Pew dataset is given in \Cref{pew-annot-interface}.

\subsection{Dataset Analysis}
\begin{figure*}[t!]
\begin{subfigure}[b]{.49\textwidth}
\centering
    \includegraphics[width=1\textwidth]{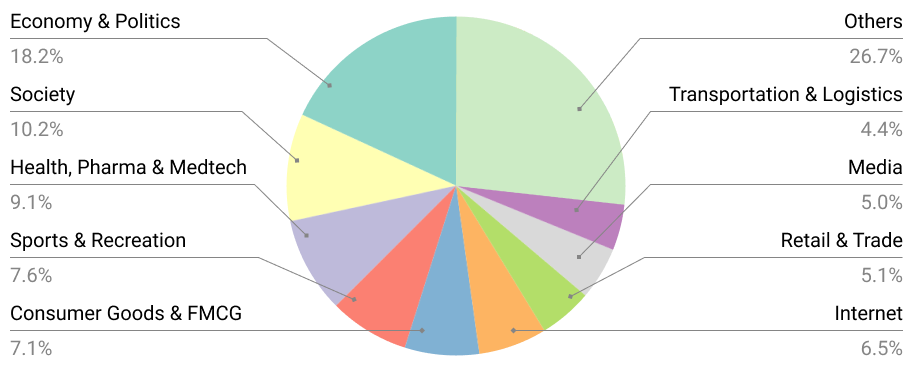}
\caption{\small Statista}
\end{subfigure}
\begin{subfigure}[b]{.49\textwidth}
\centering
    \includegraphics[width=1\textwidth]{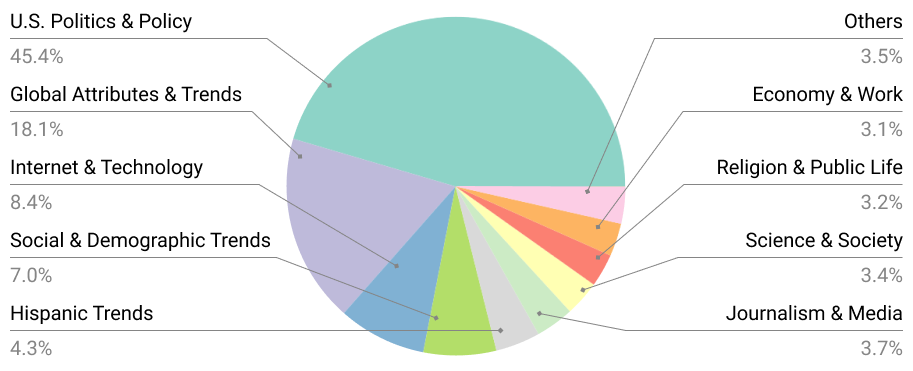}
\caption{\small Pew}
\end{subfigure}
\vspace{-0.5em}
\caption{\small Distribution of topics in the two datasets.}
\label{topics}
\end{figure*}

 Figure~\ref{topics} shows the distribution of topics in two datasets.
 
\subsection{Chart-to-text Baseline Models}
\label{app:baselines}

The experiments are done on our machine (CPU: Intel(R) Xeon(R) Gold 6240 CPU @ 2.60GHz, GPU: 4 $\times$  NVIDIA GTX 2080Ti). Training T5 is the most computationally costly task, which takes around 16-20 hours on 4$\times$ GPUs.

\paragraph{Image Captioning Models}
For pretraining the image encoders and  captioning model, we follow the same training setup as presented in the original papers. 
Inference is done with beam search with a beam size of 4.

\paragraph{Chart2text} We follow the same settings 
of \citet{obeid} with 1 encoder layer, 6 decoder layers and a dropout ratio of 0.1, and train the model for 80 epochs with a batch size of 6. For inference, we use beam search with a beam size of 4.

\paragraph{Field-Infusing Model} We follow the same settings 
as \citet{chen2020logical} and train the model for 10 epochs with a dropout ratio of 0.1 and batch size of 1. 

\paragraph{BART} We fine-tune BART-Base\footnote{\url{huggingface.co/transformers}\label{footnote:hugging}} (140M, 6-layers) 
for 500K iterations with a batch size of 4 and evaluate after every 2,000 iterations on the validation set. The initial learning rate is set to 0.0005. For inference, we use the model with the lowest validation loss and decode with a beam size of 4. 

\paragraph{T5} Similar to BART, we fine-tune T5-Base\footref{footnote:hugging} (220M, 12-layer Transformer as the encoder and decoder) for 500K iteration with a batch size of 4 and an initial learning rate of 0.0005, evaluate after every 2,000 iterations on validation set, and use {the model with best validation loss for testing. Inference is done with beam search with a beam size 4.}

\label{app:data-analysis}

\subsection{Additional Results from Evaluation}
\label{app:results}
\subsubsection{Transfer Results}
\label{app:transfer}

\begin{table}[t]
    \centering
    \scalebox{0.75}{\begin{tabular}{llc}
        \toprule
        \textbf{Pre-train Dataset} & \textbf{Fine-tune Dataset}  & \textbf{BLEU} \\
        \midrule
        Totto & Pew  & 10.66 \\
        Totto & Statista & 37.19 \\
        Pew & Statista & 37.32 \\
        Statista & Pew  & 10.73  \\
        \bottomrule
    \end{tabular}}
    \vspace{-0.5em}
    \caption{\small Results measured by BLEU for transferability based on the T5 model. }
    \label{tab:transfer}
\end{table}

{Since both Statista and Pew share some of the topics with each other, we conduct transfer experiment to verify if pretraining on one dataset could help to improve the final results on the other. In addition, since table-to-text has similarities with our task, we also experiment with pretraining on a large-scale open-domain English table-to-text dataset ToTTo \cite{parikh2020totto} before training on our datasets. We use full table for ToTTo since our task does not contain highlighted cell. Pretraining and fine-tuning use T5-based models and have the same training procedure as described in \cref{subsec:data2text}. From \Cref{tab:transfer}, we see that pretraining on other datasets only improves the final performance by a small margin. }

\subsubsection{Performance by Chart Types}
\label{app:charttype}

\begin{table}[ht]
    \centering
    \begin{tabular}{lcccc}
    \toprule
    \textbf{Chart Types} & \bf{Bar} & \bf{Line} & \bf{Pie} & \bf{Table}  \\
    \midrule
    \textbf{BLEU}     & 36.46 & 45.28 & 21.35 & 26.12 \\
    \bf{PPL} & 10.08 & 7.53 & 8.79 & 11.34\\
    \bf{CIDEr} & 4.62 & 5.59 & 3.27 & 3.67\\
    \bf{BLEURT} & 0.14 & 0.27 & -0.13 & -0.22 \\
    \bottomrule
    \end{tabular}
    \caption{Results on Statista test set \wrt chart types.}
    \label{tab:results_charttype}
\end{table}
\vspace{-0.5em}

Chart types can influence the performance of the model. We present the performance breakdown on Statista of our best model (\ie TAB-T5)  based on chart types in \Cref{tab:results_charttype}. We observe that the model is good at summarizing simple and frequent chart types (\eg line chart), whereas the model is less effective in generating informative summaries for  complex and less frequent charts (\eg pie charts) in our datasets.

\subsubsection{Human Evaluation}
\label{app:human-eval}
\begin{figure*}
    \centering
    \includegraphics[width=.9\textwidth]{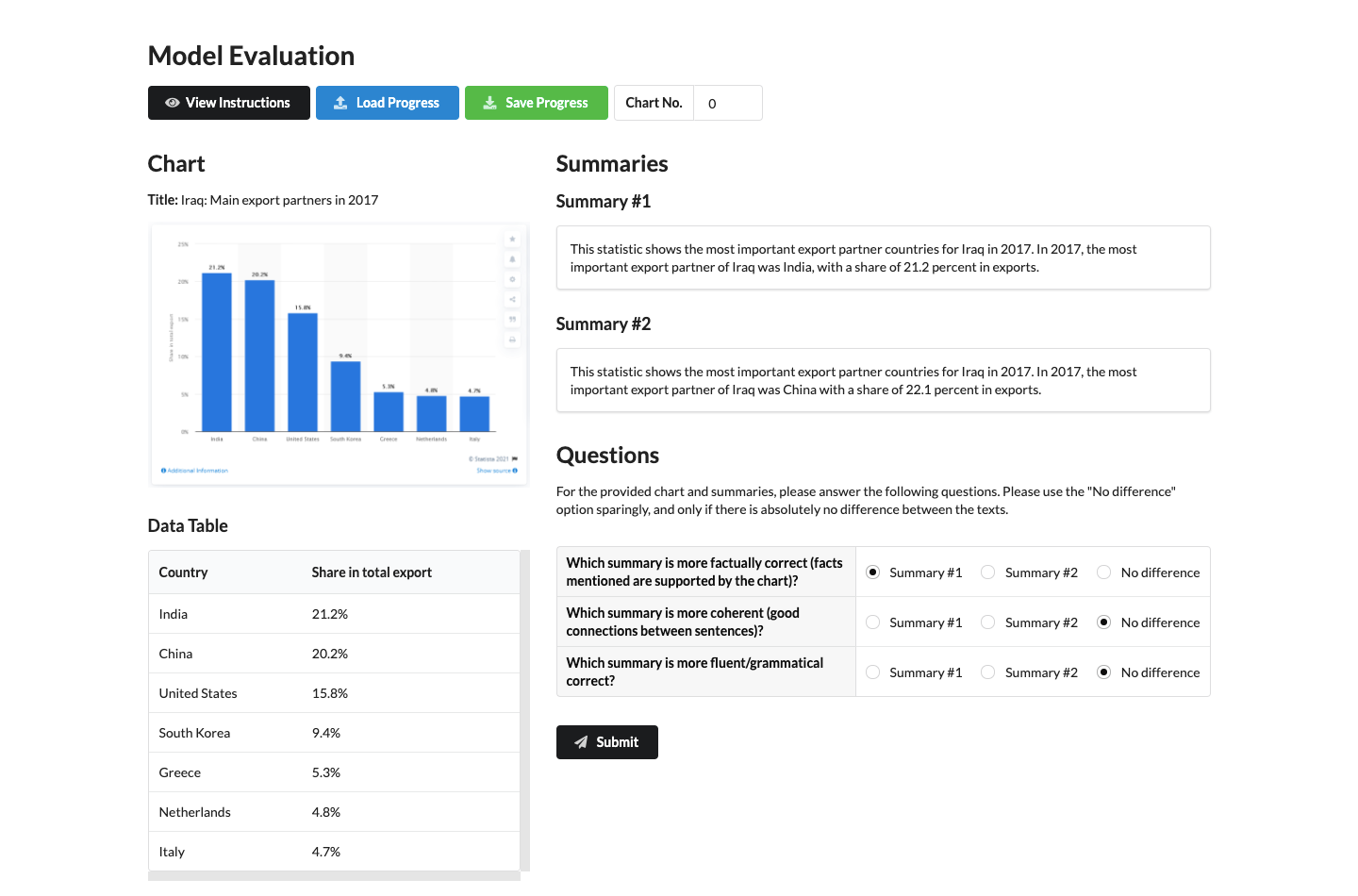}
    \caption{The user interface for human evaluation: it presents two summaries at a time and asks the participant to compare between them based on three measures.}
    \label{human-eval-interface}
\end{figure*}

The user interface for the human evaluation annotation task of comparing chart summaries is given in \Cref{human-eval-interface}. 

\begin{figure*}[t!]
    \centering
    \includegraphics[width=\textwidth]{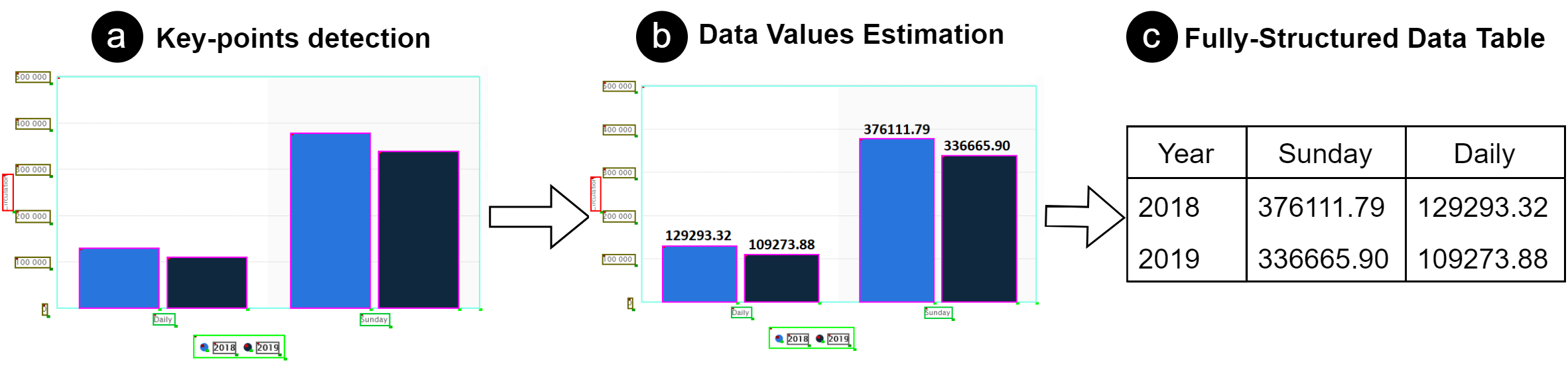}
    \caption{Data Extraction example from Statista. }
    \label{statista-data-extraction}
\end{figure*}

\subsection{Automatic Data Extraction from Charts}
\label{app:data_extraction}

\textbf{Model:} We extend ChartOCR \cite{ChartOCR} which combines deep-learning and rule-based methods to extract the underlying data values from the chart images. First, key-point detection networks detects the chart main elements (e.g. plot area, y-axis-title, x-axis-title, and legend area) and marks (e.g. bars, line points, and pie slices). We extend the detection network to detect textual labels and the legend marks in the chart (see an example in Figure \ref{statista-data-extraction}). For the rectangular objects, the network outputs the top-left and bottom-right points which are grouped together based on the distance. For lines, the network outputs the coordinates of the line points which are grouped together based on the color. For pie charts, the network outputs the separating points between the slices along the perimeter of the pie. As shown in Figure \ref{statista-data-extraction}, the scale of the chart  is estimated using the \textit{y-axis-labels'} values and y coordinates. Finally, the data values of the chart marks (e.g. bars, line points) are calculated using the scale of the chart. For pie charts, the values are estimated by calculating the angle between each two neighbouring points. 

Since the original ChartOCR model only outputs the raw data values, we we further extend their approach to output the fully-structured data table as follows. First, we utilize the CRAFT model \cite{craftrecognition} to recognize the texts of the detected textual chart elements (\textit{x-axis labels}, and \textit{legend labels}). Then, we associate the data values with their closest \textit{x-axis-label} and the data series (e.g. a group of bars or line points) with the legend labels based on the color. For example, in Figure \ref{statista-data-extraction}b, the bars are matched with their closest \textit{x-axis-labels} (`Sunday' and `Daily'). Moreover, the values of dark blue bars are associated with `2019' \textit{legend-label} and the values of light blue bars are associated with `2018' \textit{legend-label} based on the matched colors. In this way, our approach recovers the fully structured data table from the chart as shown in Figure \ref{statista-data-extraction}c.

\textbf{Evaluation Metric:} We evaluate our extracted data table using the following metric (adapted from ChartOCR \cite{ChartOCR}).
We define the distance function between two data points as:
\[
D(gt, pr) = min(1, ||\frac{gt - pr}{gt}||)
\]
where $gt$ is the ground truth value and $pr$ is the predicted value. We then compute the cost matrix $C$, where $C_{n,m} = D(gt_n, pr_m)$. The total minimum cost is then estimated by solving the linear sum assignment problem as follows:
\[
cost = \sum_{i=1}^{K}{\sum_{j=1}^{K}{C_{i, j} X_{i,j}}}
\]
Where $K = max(N, M)$ and $X$ is a binary assignment matrix. The final score is then computed using the following equation:
\[
score = 1 - \frac{cost}{K}
\]
Finally, we average the scores of all the charts to compute the overall score.

\subsection{Additional Examples from Statista and Pew datasets}
\begin{figure*}[t] 
\linespread{0.5}\selectfont\centering 
\scalebox{0.9}{\begin{tabular}{p{7.0cm} | p{8cm} }    
\toprule
        \raisebox{-1\height}{\includegraphics[height=5cm]{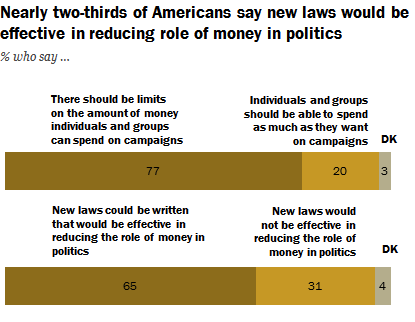}}
        &
        {Americans overwhelmingly support limits on political campaign spending, and most think new laws could effectively reduce the role of money in politics.
And there is extensive support for reining in campaign spending: 77\% of the public says “there should be limits on the amount of money individuals and organizations” can spend on political campaigns; just 20\% say they should be able to spend as much as they want.
A somewhat smaller majority (65\%) says that new campaign finance laws could be written that would be effective in reducing the role of money in politics, while 31\% say any new laws would not be effective.}
        \\
\midrule 
        \raisebox{-1\height}{\includegraphics[height=7.0cm]{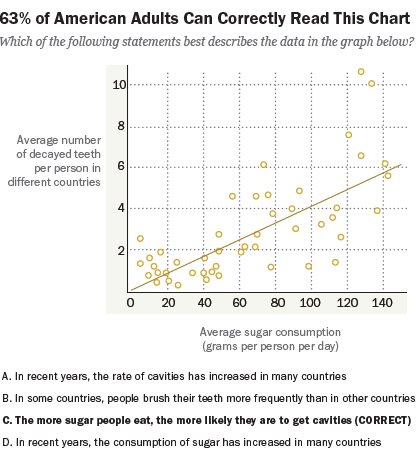}}
        & 
        {In a recent survey of what Americans know about science, we asked people to interpret the chart you see here and tell us what it showed. Six-in-ten (63\%) identify the best interpretation of this chart as “the more sugar people eat, the more likely they are to get cavities.”}
        \\
\midrule

        \raisebox{-1.0\height}{\includegraphics[height=4.0cm]{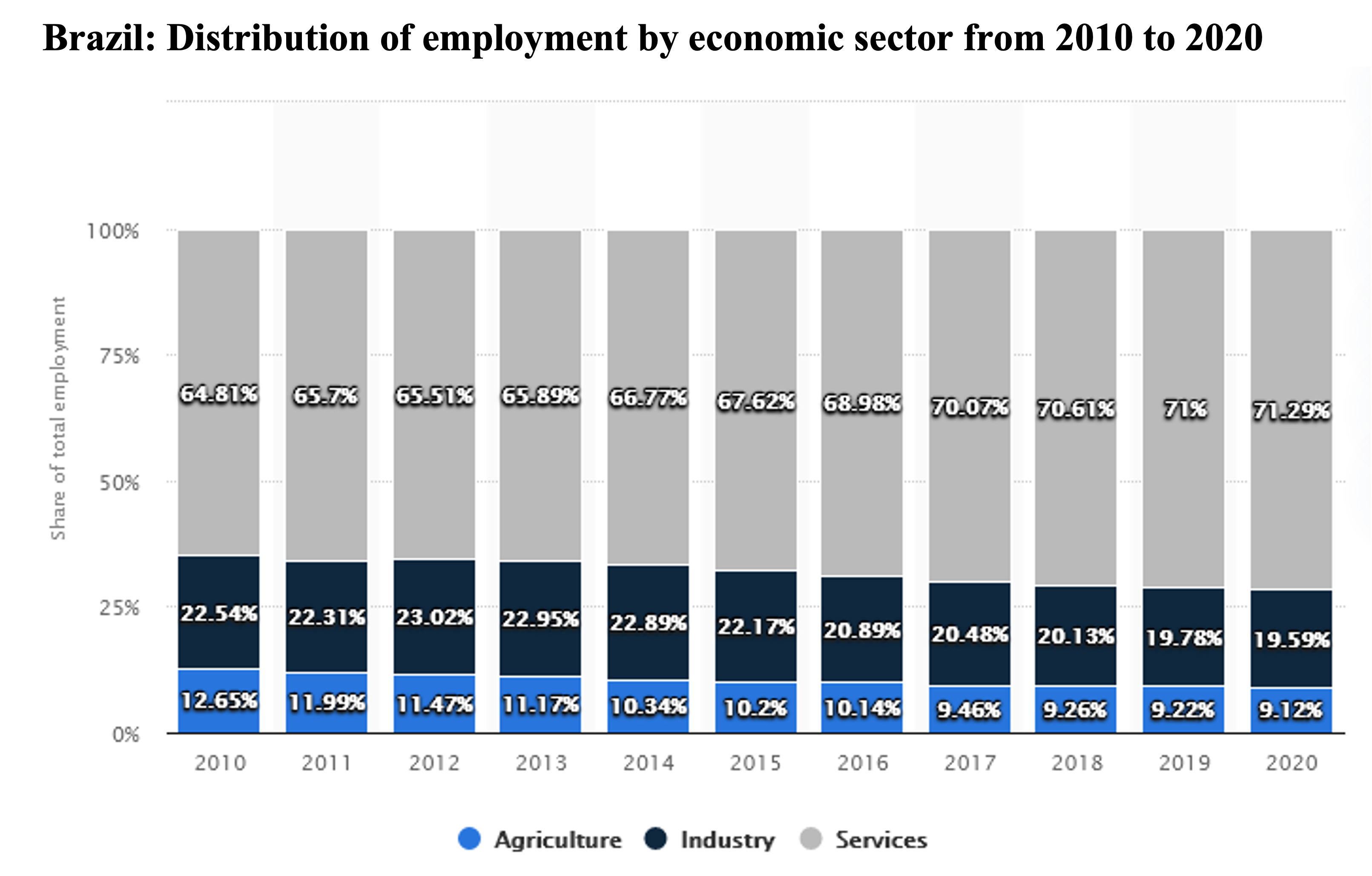}}
        &
        {The statistic shows the distribution of employment in Brazil by economic sector from 2010 to 2020. In 2020, 9.12 percent of the employees in Brazil were active in the agricultural sector, 19.59 percent in industry and 71.29 percent in the service sector.}
        \\
\midrule  
        \raisebox{-1.0\height}{\includegraphics[height=5cm]{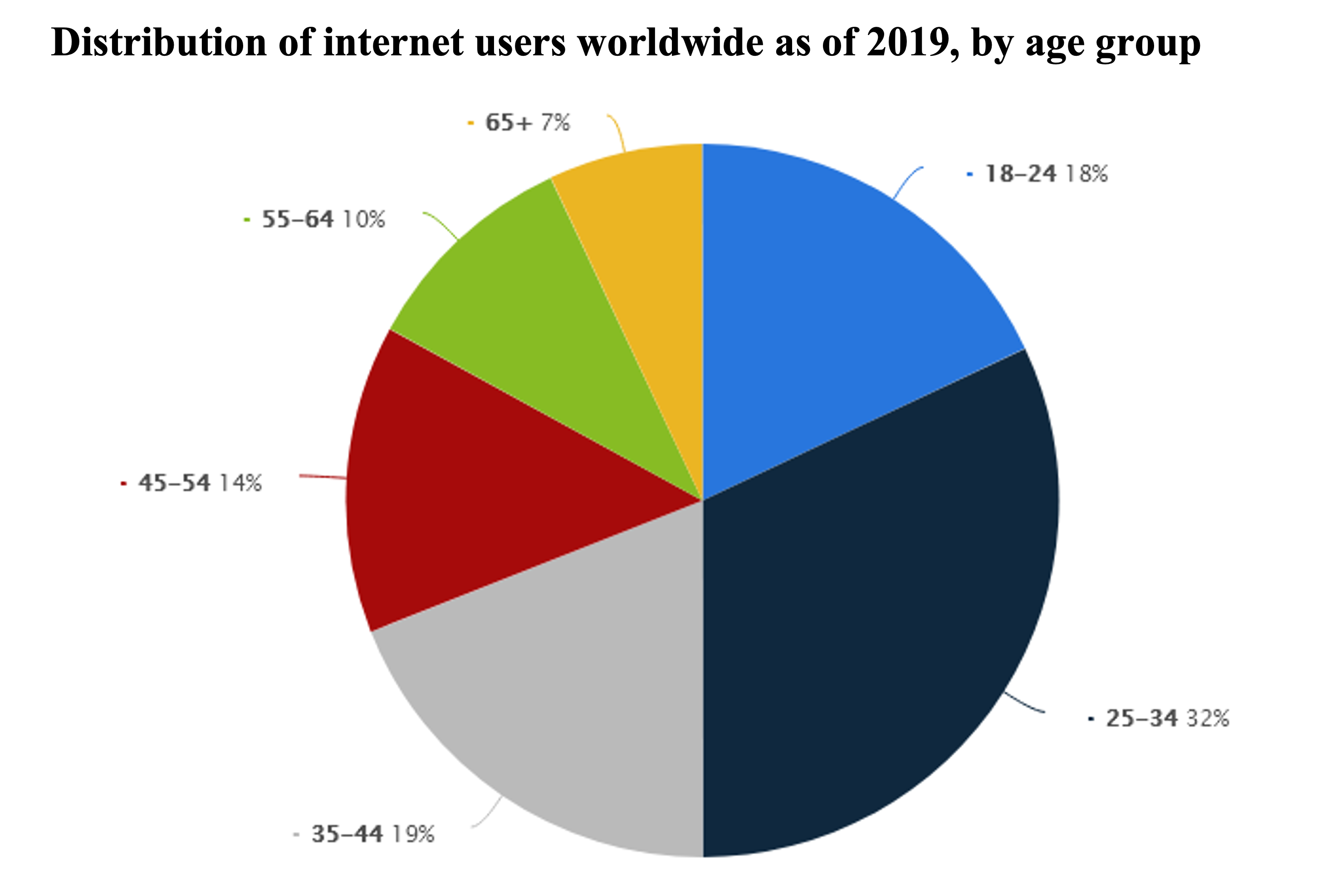}}
        &
        {As of 2019, a third of online users worldwide were aged between 25 and 34 years. Website visitors in this age bracket constituted the biggest group of online users worldwide. Also, 18 percent of global online users were aged 18 to 24 years. }
        \\
\midrule  
        
        \raisebox{-1.0\height}{\includegraphics[height=5cm]{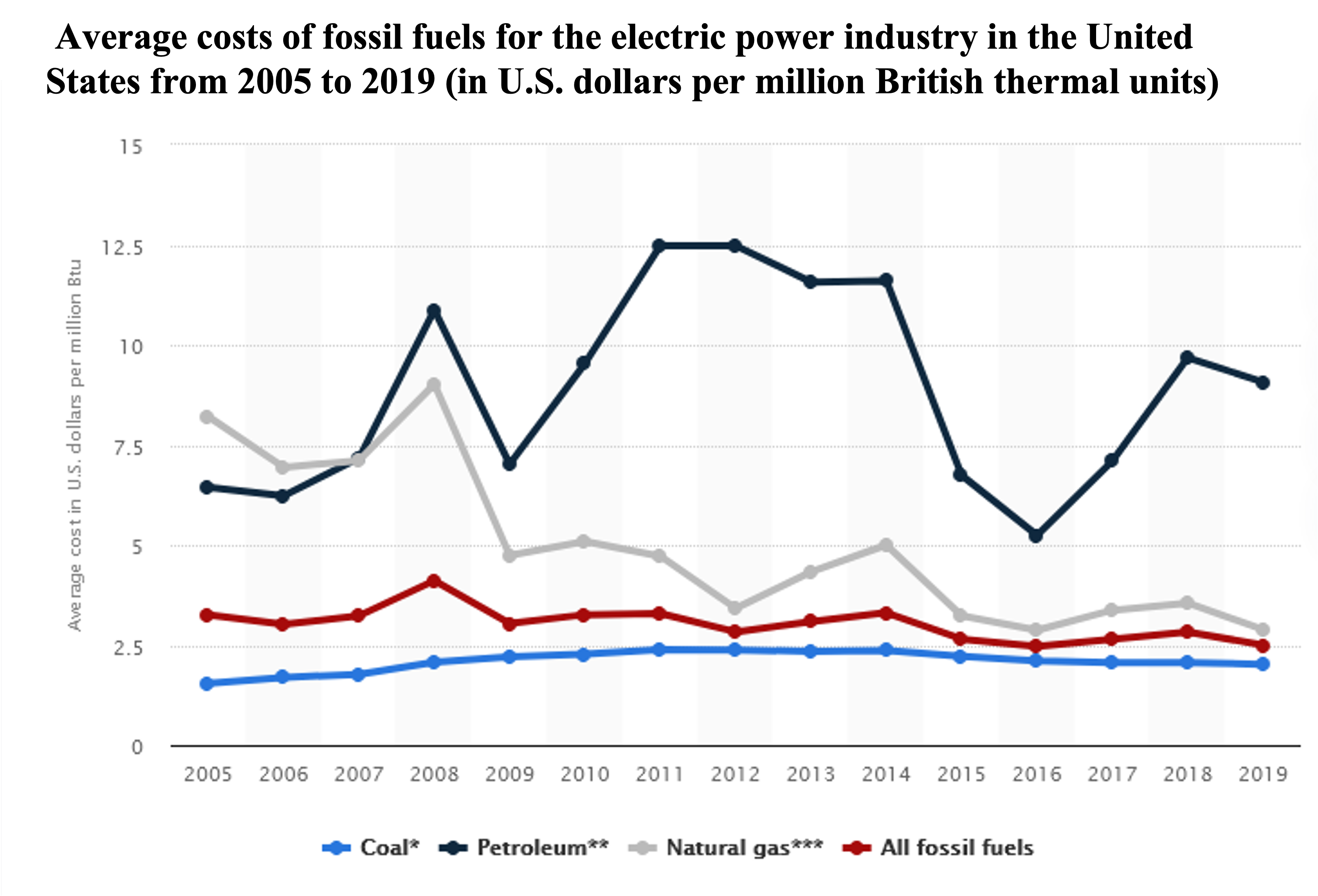}}
        &
        { The cost of fossil fuels in the electric power industry can vary depending on the source that is used. In general, fossil fuels cost about 2.50 U.S. dollars per million British thermal units (Btu) but can range from 2.02 U.S. dollars per million Btu for coal to 9.07 U.S. dollars per million Btu for petroleum. }
        \\
\bottomrule  
    \end{tabular}
    }
    \caption{Examples of chart-summary pairs in our benchmark. The top two examples are from the Pew research dataset and the rest of the examples are from the Statista dataset.
    }
    \label{tab:more examples}
\end{figure*}

Figure~\ref{tab:more examples} presents additional samples from our chart-to-text benchmark covering a diverse range of chart types and styles.

\end{document}